\documentclass[12pt]{article} %***
\usepackage[sectionbib]{natbib}
\usepackage{array,epsfig,fancyheadings,rotating}
\usepackage[]{hyperref}  %<----modified by Ivan
\usepackage{sectsty, secdot}
\sectionfont{\fontsize{12}{14pt plus.8pt minus .6pt}\selectfont}
\renewcommand{\theequation}{\thesection\arabic{equation}}
\subsectionfont{\fontsize{12}{14pt plus.8pt minus .6pt}\selectfont}
\usepackage{amsmath}
\usepackage{amssymb}
\usepackage{amsfonts}
\usepackage{multirow}
\usepackage{amsthm}
\usepackage{xcolor}
\usepackage[]{natbib}
\usepackage{dsfont}
\usepackage{enumerate}
\usepackage{algorithm}
\usepackage{algorithmic}
\usepackage{caption}
\usepackage{subcaption} % 核心：定义 subfigure 环境

\newtheorem{theorem}{Theorem}

\theoremstyle{definition}
\newtheorem{definition}{Definition}

\begin{document}

%%%%%%%%%%%%%%%%%%%%%%%%%%%%%%%%%%%%%%%%%%%%%%%%%%%%%%%%%%%%%%%%%%%%%%%%%%%%%%%%%%%%%%%%%%%%%%%%%%%%%%%%%%%%%%%%%%%%%%%%%%%%
%%%%%%%%%%%%%%%%%%%%%%%%%%%%%%%%%%%%%%%%%%%%%%%%%%%%%%%%%%%%%%%%%%%%%%%%%%%%%%%%%%%%%%%%%%%%%%%%%%%%%%%%%%%%%%%%%%%%%%%%%%%%

\renewcommand{\baselinestretch}{2}

\markright{ \hbox{\footnotesize\rm Statistica Sinica
%{\footnotesize\bf 24} (201?), 000-000
}\hfill\\[-13pt]
\hbox{\footnotesize\rm
%\href{http://dx.doi.org/10.5705/ss.20??.???}{doi:http://dx.doi.org/10.5705/ss.20??.???}
}\hfill }

\markboth{\hfill{\footnotesize\rm SUMIN WANG, CHENXIAN HUANG, YONGDAO ZHOU AND MIN--QIAN LIU} \hfill}
{\hfill {\footnotesize\rm GENERATIVE QUASI-RANDOM SAMPLING FOR COPULAS} \hfill}

\renewcommand{\thefootnote}{}
$\ $\par

%%%%%%%%%%%%%%%%%%%%%%%%%%%%%%%%%%%%%%%%%%%%%%%%%%%%%%%%%%%%%%%%%%%%%%%%%%%%%%%%%%%%%%%%%%%%%%%%%%%%%%%%%%%%%%%%%%%%%%%%%%%%

\fontsize{12}{14pt plus.8pt minus .6pt}\selectfont \vspace{0.8pc}
\centerline{\large\bf A GENERATIVE APPROACH TO QUASI-RANDOM }
\vspace{2pt} 
\centerline{\large\bf SAMPLING FROM COPULAS VIA }
\vspace{2pt}
\centerline{\large\bf  SPACE-FILLING DESIGNS}
\vspace{.4cm} 
\centerline{Sumin Wang$^1$, Chenxian Huang$^2$, Yongdao Zhou$^2$, Min-Qian Liu$^2$\footnote{\hskip-6mm Corresponding author: Min-Qian Liu, NITFID, LPMC $\&$ KLMDASR, School of Statistics and Data Science,   Nankai University, Tianjin 300071, China. E-mail: mqliu@nankai.edu.cn} } 
\vspace{.4cm} 
\centerline{\it $^1$Hebei University of Technology and $^2$ Nankai University}
 \vspace{.55cm} \fontsize{9}{11.5pt plus.8pt minus.6pt}\selectfont

%%%%%%%%%%%%%%%%%%%%%%%%%%%%%%%%%%%%%%%%%%%%%%%%%%%%%%%%%%%%%%%%%%%%%%%%%%%%%%%%%%%%%%%%%%%%%%%%%%%%%%%%%%%%%%%%%%%%%%%%%%%%

\begin{quotation}
\noindent {\it Abstract:}
{\bf }\\
Exploring the dependence between covariates across distributions is crucial for many applications. Copulas serve as a powerful tool for modeling joint variable dependencies and have been effectively applied in various practical contexts due to their intuitive properties. However, existing computational methods lack the capability for feasible inference and sampling of any copula, preventing their widespread use. This paper introduces an innovative quasi-random sampling approach for copulas, utilizing generative adversarial networks (GANs) and space-filling designs. The proposed framework constructs a direct mapping from low-dimensional uniform distributions to high-dimensional copula structures using GANs, and generates quasi-random samples for any copula structure from points set of space-filling designs. In the high-dimensional situations with limited data, the proposed approach significantly enhances sampling accuracy and computational efficiency compared to existing methods. Additionally, we develop convergence rate theory for quasi-Monte Carlo  estimators, providing rigorous upper bounds for bias and variance. Both simulated experiments and practical implementations, particularly in risk management, validate the proposed method and showcase its superiority over existing alternatives.

\vspace{9pt}
\par
\end{quotation}\par

\def\thefigure{\arabic{figure}}
\def\thetable{\arabic{table}}

\renewcommand{\theequation}{\thesection.\arabic{equation}}

\fontsize{12}{14pt plus.8pt minus .6pt}\selectfont

\section{Introduction}\label{sec-intro}

Modeling multivariate dependencies  is a central task in statistical analysis, as it aids in comprehending the complex dependence between covariates  across the entire distribution.  For example, in machine learning, a major interest lies in enforcing spatial dependence in convolutional neural networks or temporal dependence in recurrent neural networks \citep{ng2022}. Copulas are a tool for modeling the joint dependency of random variables by separating the modeling of univariate marginals from the complex multivariate structure \citep{joe2014}.  This is an important property in fields such as financial applications. Besides practical applications, quasi-random sampling for copulas is also interesting from a theoretical perspective \citep{hofert2008sampling}. This is particularly significant  in a variety of  problems. For examples, in numerical integration, quasi-random sampling can enhance the classical Monte Carlo (MC) method by replacing pseudo-random samples \citep{cambou2017}.

\fancyhf{} %
\setlength\headheight{23pt}
\chead[]{{\footnotesize\rm GENERATIVE QUASI-RANDOM SAMPLING FOR COPULAS}}
\fancyhead[RE,RO]{\thepage}

Generating quasi-random samples for copulas is generally a non-trivial task. For independent copulas, one can employ randomized quasi-Monte Carlo (QMC) techniques, such as low-discrepancy sequences \citep{lemieux2009} or space-filling designs in computer experiments  \citep{Santner2018}. In contrast, for dependent copulas, many  authors \citep[e.g.,][]{zhu2014,aistleitner2015functions} have utilized the inverse Rosenblatt transform \citep{rosenblatt1952remarks}, a method known as the conditional distribution method \citep[CDM,][]{embrechts2001modelling,hofert2010sampling}. The objective of CDM is to identify an optimal transport map, $\phi_C$, which maps randomized   QMC points from the uniform distribution $U[0,1]^k$ to samples distributed according to a target $d$-dimensional copula $C$. A prerequisite for this mapping is typically $k \geq d$.  As highlighted by \cite{cambou2017},  closed-form expressions for the inverse Rosenblatt transform are known for certain copulas, such as the normal, $t$, and Clayton families. However, for most other copulas, such analytical solutions are not available, making CDM computationally demanding.
  In other words, there is no universally applicable and numerically efficient approach to generate the desired copula's quasi-random samples.   The key challenge lies in devising a computationally efficient method for constructing  the transformation $\phi_C$.

  In recent years, there has been a significant increase in interest in utilizing generative models for copula learning. For instance, \cite{ling2020} proposed a differentiable neural network architecture for learning the dependency structure of Archimedean copulas. Similarly, \cite{janke2021} introduced a conditional bivariate copula hierarchical modeling approach based on implicit generative networks, which addresses the restrictive assumptions regarding marginal distributions prevalent in conventional methods. \cite{ng2022} pioneered the integration of deep generative models, including GANs and variational autoencoders, with Archimax copulas, proposing a highly scalable non-parametric inference and sampling framework that exhibits remarkable performance in modeling tail dependencies. Furthermore, \cite{Pascal2024} introduced a hybrid framework that integrates Bayesian networks and variational autoencoders to jointly learn the dependency structures and marginal distribution characteristics of copulas, achieving high-fidelity synthesis of population data. In the field of the Internet of Things, \cite{ragothaman2025access} proposed a copula structure learning framework based on GANs. This framework employs deep neural networks to model the joint distribution characteristics of multidimensional policy variables, and then utilizes inverse transform sampling to generate access control policy data that complies with real-world constraints derived from uniform distributions.

However, the literature on using generative models for quasi-random sampling of copulas remains relatively underdeveloped. \cite{hofert2021}  was a pioneer in this field, introducing using generative moment matching networks \citep[GMMN,][]{li2015generative} as a generative approach for quasi-random sampling of multivariate distributions. They employed a generative neural network to approximate the transformation  $\phi_C$ and trained it using maximum mean discrepancy optimization. This computational method enables the generation of approximate quasi-random samples from any copula. Despite its potential, GMMN has been found to be less robust \citep{janke2021} and faces limitations. For example, \cite{li2017} pointed out that GMMN may not perform as competitively as generative adversarial networks (GANs) on complex and large-scale datasets. Additionally, GMMN is generally less computationally efficient due to its reliance on large batch sizes during training. Although GMMN-based quasi-random sampling techniques have wide applications, the theoretical properties  of quasi-randomness have yet to be fully explored.

To address these limitations, we propose an efficient quasi-random sampling method that is computationally efficient and enjoys nice theoretical properties. The proposed  method employs GANs to learn the optimal transformation $\phi_C$, followed by the use of space-filling designs to generate randomized QMC points on low-dimensional uniform distributions. These points are then mapped to the copula  via $\phi_C$.  Essentially, the proposed method is an optimal transport method \citep{fang2025spot}, and shares similarities with the framework of optimal sampling in an embedding space \citep{zhang2022optimal}.
 The proposed method offers several advantages: (a) $\phi_C$ in the method represents the optimal mapping between low-dimensional uniformity and high-dimensional distributions, bypassing the complexities of high-dimensional space-filling designs. (b) The method outperforms traditional techniques like CDM, which often rely on stringent parameter assumptions and are less effective in high-dimensional settings.   (c) The method yields a more accurate $\phi_C$ compared to GMMN when data are limited, and this advantage persists when  $k<d$.  Therefore, it exhibits lower variance and stronger robustness across $d=10,20,200$ (see Section~5).  (d) A comprehensive analysis of the theoretical properties of quasi-randomness specifically for low-variance  MC estimators is presented,  and  upper bounds on the bias and variance of these estimators are established.

The rest of the paper is organized as follows. Section 2 introduces some  preliminaries  including copulas, and space-filling designs (including low-discrepancy sequences, Latin hypercube designs and their variants), as well as GANs.  The proposed method and related theoretical analysis are presented in  Section 3. Section 4 shows a series of numerical simulations. Section 5 presents a real data analysis in risk management.  We conclude this paper in Section 6. All the proofs are provided in the supplementary materials.

\section{ Preliminaries}

 \subsection{ Problem Setting and Copulas}
As widely recognized, the primary application of quasi-random sampling lies in obtaining low-variance MC estimators. For instance, many problems in financial derivative pricing and Bayesian computation can be reduced to the computations of expectations. The quantity of interest is
 $\mu=E\left[\Psi_0(\boldsymbol{X})\right]$, where $\boldsymbol{X}=\left(X_1, \ldots, X_d\right): \Omega \rightarrow \mathbf{R}^d$ is a random vector with  distribution function $F$ on a probability space $(\Omega, \mathcal{F}, P)$, and $\Psi_0: \mathbf{R}^d \rightarrow \mathbf{R}$ is a measurable function. The components of $\boldsymbol{X}$ are typically dependent, and it is common to employ a  copula $C$ to model the joint distribution function $F$.    This relationship can be described using Sklar's Theorem \citep{nelsen2006,joe2014},  which asserts that $F$ can be represented as a composition of the marginals and the copula, connecting the dependence structure among the variables, i.e.
\begin{equation}\label{eq1}
 F(\boldsymbol{x})=C\left(F_1\left(x_1\right), \ldots, F_d\left(x_d\right)\right), \quad \boldsymbol{x}=(x_1,\ldots, x_d) \in \mathbf{R}^d,
 \end{equation}
where $F_j(x)=P\left(X_j \leq x\right)$ for $j \in\{1, \ldots, d\}$ are the marginal distribution functions of $F$. $C:[0,1]^d \rightarrow[0,1]$ is the unique underlying copula, which is a distribution function with standard uniform univariate margins. A copula model, such as the one described in~\eqref{eq1}, allows for the separation of the dependence structure from the marginal distributions. This is particularly valuable when considering model building and sampling, especially in cases where $E\left[\Psi_0(\mathbf{X})\right]$ primarily depends on the dependence between the components of $\mathbf{X}$.

In terms of the copula model \eqref{eq1}, we can express
$$
\mu=E\left[\Psi_0(\mathbf{X})\right]=E[\Psi(\boldsymbol{u})],
$$
where $\boldsymbol{u} = (u_1, \ldots, u_d): \Omega \rightarrow \mathbf{R}^d$ is a random vector with the cumulative distribution function $C$, and $\Psi:[0,1]^d \rightarrow \mathbf{R}$ is defined as
$$
\Psi\left(u_1, \ldots, u_d\right)=\Psi_0\left(F_1^{-}\left(u_1\right), \ldots, F_d^{-}\left(u_d\right)\right),
$$
and $F_j^{-}(p)=\inf \left\{x \in \mathbf{R}: F_j(x) \geq p\right\}$, for $j \in\{1, \ldots, d\}$. In general,  an analytical expression for the quantity of interest $E\left[\Psi_0(\boldsymbol{X})\right]$ rarely exists, and thus numerical methods must be applied to evaluate it.  Given a dataset  $\{\boldsymbol{X}_i\}_{i=1}^{N}$  sampled from distribution  $F$, after employing established techniques to estimate the copula  $C$ and marginal distribution $F_j$, $j \in \{1, \ldots, d\}$,  MC simulation can be employed to approximate  $E[\Psi(\boldsymbol{u})]$. One advantage of MC simulation is that the rate of convergence of its error is independent of the dimensionality of a given problem. Nevertheless, the convergence rate of plain MC is generally slow,  so that MC is often combined with some variance reduction technique  to improve the precision of estimators.

A primary challenge in MC simulations is the efficient sampling from copulas. To enhance this process, QMC simulation, a well-established variance reduction technique \citep{niederreiter1992, owen2008}, is often employed. QMC methods typically achieve faster convergence rates than traditional MC by replacing pseudo-random numbers with quasi-random samples in the sampling algorithm. These quasi-random samples, generated from low-discrepancy sequences or space-filling designs, are constructed to provide more uniform coverage of the probability space. This superior uniformity makes QMC particularly effective for complex numerical integration, leading to its widespread adoption in fields such as finance \citep{paskov1996}, option pricing \citep{he2023}, and fluid mechanics \citep{graham2011, kuo2012}. Accordingly, the focus of this paper is on the generation of quasi-random samples from a target copula $C$ corresponding to a distribution $F$.

\subsection{Discrepancy and Space-Filling Designs}

To generate  quasi-random samples of $C $,  we first need to compute the transformation $\phi_C$.  Once $\phi_C$ is obtained, we can use some space-filling designs to create the randomized QMC  points  on the unit hypercube $[0,1] ^ k $. These points are then mapped to the desired quasi-random samples of $ C $ through $ \phi_C $.
Our method heavily relies on space-filling designs, which are important in computer experiments. These designs, unlike random sampling,  are not to mimic i.i.d. samples, but rather to achieve a homogeneous coverage of $[0,1]^k$. Two popular methods for achieving this are low-discrepancy sequences (or uniform designs) and Latin hypercube designs (LHDs),  which  are assessed using discrepancy criteria and stratification properties.
 %which  are assessed using discrepancy criteria  ?and stratification properties.

%In our method, space-filling design plays a crucial role. Therefore, we will provide a brief introduction about  space-filling design.
%Space-filling design, a general strategy for computer experiments, aim to distribute points uniformly across the entire hypercube as evenly as possible, avoiding sparse or  regions compared to random sampling. The objective is not to mimic independent and identically distributed (i.i.d.) samples but to ensure homogeneous coverage of the entire space. Homogeneity can be assessed using discrepancy criteria and stratified characteristics, resulting in
%two commonly used  points in $U [0,1]^k$, i.e. low-discrepancy sequence (or uniform design) and Latin hypercube design.

 The low-discrepancy sequence is developed upon the notion of star discrepancy, which is a classical metric that measures the discrepancy between a set of discrete data points and the uniform distribution on the unit hypercube $[0,1]^k$.  Let $P_{n}=\left\{\boldsymbol{v}_1, \ldots, \boldsymbol{v}_{n}\right\}$ be a set of $n$ data points in $U[0,1]^k$, and  $[\boldsymbol{0},\boldsymbol{a})=\prod_{i=1}^{k}[0, a_i)$ be a hyper-rectangle, where $\boldsymbol{0}=(0,\ldots, 0)$, and $\boldsymbol{a}=(a_1,\ldots, a_k) \in [0,1]^k$. Then the star discrepancy of $P_{n}$ is  defined as follows.

 \begin{definition} \rm {
   Given $P_{n}$ and a hyper-rectangle $[\mathbf{0}, \boldsymbol{a})$, the corresponding local discrepancy is defined as, $D\left(P_{n}, \boldsymbol{a}\right)=$ $\left|\frac{1}{n} \sum_{i=1}^n \mathds{1}\left\{\boldsymbol{v}_i \in[\mathbf{0}, \boldsymbol{a})\right\}-\prod_{j=1}^k a_j\right|$. Here, $ \mathds{1}\left\{\boldsymbol{v}_i \in[\mathbf{0}, \boldsymbol{a})\right\}$ denotes the indicator function, which equals $1$ if the  point $\boldsymbol{v}_i$ falls in $[\mathbf{0}, \boldsymbol{a})$,  and $0$ otherwise. The star discrepancy is defined as
$$
D^*\left(P_{n}\right)=\sup _{\boldsymbol{a} \in[0,1]^k} D\left(P_{n}, \boldsymbol{a}\right) .
$$
}
 \end{definition}

There exist many methods that generate design points via directly minimizing the star discrepancy, and these methods are called  low-discrepancy sequences,  such as Sobol sequences or  generalized Halton sequences \citep{lemieux2009}, and uniform design methods \citep{fang2018}.

Another type of space-filling design  based on stratification properties  is the LHD  \citep{Mckay1979}. An LHD exhibits a key characteristic of one-dimensional uniformity, ensuring that each input variable is uniformly distributed across its range, which is defined as follows.
%In the following sections, we will first introduce the definition of an orthogonal array and then explain how to generate a random LHS based on an orthogonal array.
\begin{definition}
\rm {
  An  LHD, denoted by $D=(d_{ij})_{n\times k}$,  is an $n \times k$ matrix constructed by
$$
d_{i j}=\pi_j(i) / n+\eta_i^j / n, ~~ i=1,\ldots,n, ~~ j=1,\ldots, k,
$$
where the $\pi_j$'s are uniform permutations on $\{0,1, \ldots, n-1\}$, and  $\eta_i^j$'s are generated independently from uniform distributions on $[0,1]$. These permutations $\pi_j$'s and random variables   $\eta_i^j$'s are generated independently.
}
\end{definition}

 To achieve multi-dimensional space-filling, recent studies have employed orthogonal array (OA)-based LHD  \citep{Tang1993,ai2016}. This methodology begins with an orthogonal array, as defined in Definition \ref{oa}, and subsequently constructs a random Latin hypercube design through level-wise expansion.
\begin{definition}\label{oa} {\rm \citep{hedaya1999}.}
{\rm
An orthogonal array with $n$ rows, $k$ columns, and strength $t$ (where $1 \leq t \leq k$), denoted by $OA(n, s^k, t)$, is an $n \times k$ matrix where each column consists of $s$ levels drawn from the set $\{0,1,\ldots,s-1\}$, and all possible level combinations occur equally often as rows in every $n \times t$ submatrix.}
\end{definition}

%Let $Z_n$ denote the set $\{1, \ldots, n\}$ for any positive integer $n$.
Let $ A$ be an $O A\left(n, s^k, t\right)$, \cite{Tang1993} proposed  a random  OA-based LHD $D=(d_{ij})_{n \times k}$ based on $A$, which is described in the following steps,  and can be generated using the R package \texttt{LHD}.

 %Given an $O A\left(n, s^k, t\right)$ denoted as $A$, the process of generating a random OA-based LHD $D=(d_{ij})_{n \times k}$ consists of the following steps:
\begin{description}
 % \item [\rm{Step 1.}]  For each column of $A$, relabel the $s$ levels with a random permutation of $Z_{s}$.
  \item [\rm{Step 1.}] For $j=1, \ldots, k$ and $e=0, \ldots, s-1$, replace the $n/s$ positions of $e$ in the $j$th column of $A$ with a random permutation of $\{1,\ldots, n/s\}$. Denote by $B=\left(b_{i j}\right)_{n \times k}$ the resulting array from $A$ after such replacements.
  \item [\rm{Step 2.}] For $i=1, \ldots, n$ and $j=1, \ldots, k$, let
$$
d_{i j}=a_{i j}/s+\left(b_{i j}-\varepsilon_{i j}\right)/n,
$$
 where $a_{i j}$ is the $(i, j)$-th entry of $A$ and $\varepsilon_{i j}$'s are independent random variables following $U[0,1]$.
\end{description}

\subsection{Generative Adversarial Networks}
The challenge in the CDM method lies in computing the transformation $\phi_C$, particularly in high-dimensional copulas. To address this challenge, we propose using machine learning method, specifically generative adversarial networks \citep[GANs,][]{Goodfellow2014}, for non-parametric estimation of $\phi_C$. Given a dataset $\{\boldsymbol{X}_i\}_{i=1}^{N}$ with $\boldsymbol{X}_i=(X_{i,1}, \ldots, X_{i,d})$, our objective is to learn a transport map $\phi_C: U[0,1]^k \to C \in \mathbf{R}^d$  by GANs. Once trained, quasi-random samples from the copula $C$ are obtained by pushing randomized QMC points in $U[0,1]^k$ through $\phi_C$, where the latent dimension $k$ is allowed to be smaller than the output dimension $d$.
In the following of this subsection, we provide a brief overview of GANs.

 In this paper, we employ deep neural networks   for fitting the transformation $\phi_C$. Let $L$ be the number of (hidden) layers in the neural networks and, for each $l=0, \ldots, L+1$, let $N_l$ be the dimension of layer $l$, that is, the number of neurons in layer $l$. In this notation, layer $l=0$ refers to the input layer which consists of the input $\boldsymbol{z} \in \mathbf{R}^k$ for $N_0=k$, and layer $l=L+1$ refers to the output layer which consists of the output $\boldsymbol{x} \in \mathbf{R}^d$ for $N_{L+1}=d$. Layers $l=1, \ldots, L+1$ can be described in terms of the output $\boldsymbol{a}_{l-1} \in \mathbf{R}^{N_{l-1}}$ of layer $l-1$ via
$$
\begin{aligned}
\boldsymbol{a}_0 & =\boldsymbol{z} \in \mathbf{R}^{N_0}, \\
\boldsymbol{a}_l & =T_l\left(\boldsymbol{a}_{l-1}\right)=\sigma_l\left(A_l \boldsymbol{a}_{l-1}+\boldsymbol{b}_l\right) \in \mathbf{R}^{N_l} ~\text{for} ~ l=1, \ldots, L+1,~ \text{and} \\
\boldsymbol{y} & =\boldsymbol{a}_{L+1} \in \mathbf{R}^{N_{L+1}}
\end{aligned}
$$
with weight matrices $A_l \in \mathbf{R}^{N_l \times N_{l-1}}$, bias vectors $\boldsymbol{b}_l \in \mathbf{R}^{N_l}$, and activation functions $\sigma_l$; note that for vector inputs, the activation function $\phi_l$ is understood to be applied componentwise. Then, the neural network $\mathcal{N} \mathcal{N}(W, L): \mathbf{R}^k \to \mathbf{R}^d$ can then be written as the composition
%$$
%T_{\boldsymbol{\theta}}:=T_{L+1}\left(\sigma_{L}\left(T_{L}\left(\cdots \sigma_2\left(T_1(\boldsymbol{a}_0)\right) \cdots\right)\right)\right)
%$$
$$
T_{\boldsymbol{\theta}}:=T_{L+1}\circ  T_{L} \circ  \cdots \circ T_2 \circ T_1
$$
with its parameter vector given by $\boldsymbol{\theta} = \{(A_i, \boldsymbol{b}_i)\}_{i=1}^{L+1}$. The width of this neural network is defined as $W=$ $\max \left\{N_1, \ldots, N_L\right\}$, and if the maximum norm $\| T_{\boldsymbol{\theta}} \|_{\infty} \leq B$, we use $\mathcal{N} \mathcal{N}(W, L, B)$ to denote the neural network $T_{\boldsymbol{\theta}}$. Here,  for any function $f(\boldsymbol{x}): \mathcal{X} \rightarrow$ $\mathbf{R}^d$, denote $\|f\|_{\infty}=\sup _{x \in \mathcal{X}}\|f(x)\|$, where $\|\cdot\|$ is the Euclidean norm.

GANs  are a learning technique for high-dimensional data distributions. In GANs, adversarial learning is employed, fostering a competitive dynamic between a generator network \( G \) and a discriminator network \( D \), with the goal of producing high-quality samples. Specifically, to approximate a target distribution \( \gamma \), GANs start by sampling a vector \( \boldsymbol{z} \) from a simple source distribution $\nu$, often uniform or Gaussian, and then train the generator \( G \) to make \( G(\boldsymbol{z}) \) closely resemble \( \gamma \). The generator is derived through solving the following minimax optimization problem:
\begin{equation*}
\min_{G\in \mathcal{G}} \max_{D \in \mathcal{D}} \left( E_{\boldsymbol{x} \sim \gamma}[\log D(\boldsymbol{x})] + E_{\boldsymbol{z} \sim \nu}[1-\log D(G(\boldsymbol{z}))] \right),
\end{equation*}
where both the generator class $\mathcal{G}$ and the discriminator class $\mathcal{D}$ are  commonly parameterized using neural networks.

\section{Methodology}
\label{sec:meth}

\subsection{  Quasi-Random Sampling for Copulas Using GANs and Space-Filling Designs}
In this subsection, we  train a generator $G$ mapping samples $\mathbf{z} \sim \nu$ to samples from the copula $C$, where  $\nu$ is an easily-sampled source distribution. In practice, the source distribution $\nu$ is commonly chosen as the Gaussian distribution $N(0, I)$ with the cumulative distribution function $F_{\mathbf{z}}$. Once the generator is trained, the copula transformation $\phi_C$ can be approximated by the composition $G \circ F_{\mathbf{z}}^{-1}$. To train $G$ using GANs, the loss function is defined as follows:
 \begin{equation*}
  \mathcal{L}(G,D)=E_{\boldsymbol{u}\sim C}[\log(D(\boldsymbol{u}))]+E_{\boldsymbol{z}\sim \nu}[\log(1-D(G(\boldsymbol{z})))].
\end{equation*}
  The target  generator $G^*$ and the target discriminator $D^*$ are characterized by the minimax optimization problem:
\begin{equation*}
\left(G^*, D^*\right)=\operatorname{argmin}_G \operatorname{argmax}_D \mathcal{L}(G, D).
\end{equation*}

 In fact, the precise form of the distribution $C$ remains unknown; we only have a set of observed samples $\boldsymbol{X}_1, \ldots, \boldsymbol{X}_N$. According to \cite{hofert2018b}, we can generate pseudo-samples $\{\boldsymbol{u}_i\}_{i=1}^{N}$ that follow $C$ by using the method in (\ref{psu}).
  \begin{equation}\label{psu}
\boldsymbol{u}_{i}=\frac{1}{N+1}\left(R_{i 1}, \ldots, R_{i d}\right), \quad i =1, \ldots, N,
\end{equation}
where
$R_{ij}$  denotes the rank of element $X_{ij}$ among $X_{1j},\ldots, X_{Nj}$. These synthetic samples can be effectively employed to facilitate the training of the generator $G^*$.  Assume we also have  $N$ i.i.d. samples  $\{\boldsymbol{z}_j\}_{j=1}^{N}$ drawn from the source distribution  $\nu$.  We now consider the empirical version of the loss function $\mathcal{L}(G,D)$, given by:
\begin{equation}\label{emloss}
\widehat{\mathcal{L}}(G, D)=\frac{1}{N} \sum_{i=1}^N \log(D(\boldsymbol{u}_i))+\frac{1}{N} \sum_{i=1}^N [\log(1-D(G(\boldsymbol{z}_i)))].
\end{equation}

Throughout the paper, we estimate the generator  $G$ and discriminator  $D$ using nonparametric neural networks.
Specifically, we employ two feedforward neural networks: the generator network $G_{\boldsymbol{\theta}}$ with parameters $\boldsymbol{\theta} \in \mathcal{G} = \mathcal{NN}(L_1, W_1, B_1)$ is a Softplus-activated neural network mapping $\mathbb{R}^k \to [0,1]^d$, where $L_1$ is the depth, $W_1$ is the width, and $B_1$ is the maximum norm constraint $\left\|G_{\boldsymbol{\theta}}\right\|_{\infty} \leq B_1$. The Softplus activation is defined as $\sigma(\boldsymbol{x}) := \log(1+e^{\boldsymbol{x}})$.
%the generator network $G_{\boldsymbol{\theta}}$ with parameter $\boldsymbol{\theta}$, belonging to the set $\mathcal{G} = \mathcal{NN}(L_1, W_1, B_1)$, which employs the ReLU active function based networks mapping from $\mathbf{R}^k$ to $[0,1]^d$ with depth $L_1$, width $W_1$, and a maximum norm constraint $\left\|G_{\boldsymbol{\theta}}\right\|_{\infty} \leq B_1$. Here the ReLU active function is denoted by $\sigma (\boldsymbol{x}):=\boldsymbol{x} \vee \boldsymbol{0}$.
Similarly, the discriminator network $D_{\boldsymbol{\phi}}: [0,1]^d \rightarrow \mathbf{R}$ has parameter $\boldsymbol{\phi}$  belonging to the set $\mathcal{D} = \mathcal{NN}(L_2, W_2, B_2)$, with depth $L_2$, width $W_2$, and a norm bound $\left\|D_{\boldsymbol{\phi}}\right\|_{\infty} \leq B_2$.
The estimation of $\boldsymbol{\theta}$ and $\boldsymbol{\phi}$ is achieved by solving the empirical form of the minimax optimization problem.
$$
(\hat{\boldsymbol{\theta}}, \hat{\boldsymbol{\phi}})=\operatorname{argmin}_{\boldsymbol{\theta}} \operatorname{argmax}_{\boldsymbol{\phi}} \widehat{\mathcal{L}}\left(G_{\boldsymbol{\theta}}, D_{\boldsymbol{\phi}}\right) .
$$

We train the discriminator and generator in an iterative manner, updating the parameters $\boldsymbol{\theta}$ and $\boldsymbol{\phi}$ alternately, as follows:
\begin{enumerate}[(a)]
  \item  Fix $\boldsymbol{\theta}$, update the discriminator by ascending the stochastic gradient of the loss (\ref{emloss}) with respect to $\boldsymbol{\phi}$.
  \item  Fix $\boldsymbol{\phi}$, and update the generator by descending the stochastic gradient of the loss (\ref{emloss}) with respect to $\boldsymbol{\theta}$.
\end{enumerate}

The detailed training procedure is outlined in Algorithm \ref{alg2}, which employs pseudo-samples $\{\boldsymbol{u}_{i}\}_{i=1}^{N}$. We denote the estimated generator as $\hat{G} = G_{\hat{\boldsymbol{\theta}}}$ and the estimated discriminator as $\hat{D} = D_{\hat{\boldsymbol{\phi}}}$, where $\hat{\boldsymbol{\theta}}$ and $\hat{\boldsymbol{\phi}}$ represent the learned parameters for $G$ and $D$, respectively.

%Next, we describe the process of obtaining quasi-random samples of the copula $C$ using the generator network $\hat{G}$. This process involves two steps: obtaining quasi-random samples from $U[0,1]^k$ and then transforming them into quasi-random samples of $C$ through $G_{\hat{\theta}}\circ F_{\boldsymbol{z}}^{-1}$. To generate a desired number of quasi-random samples $\{\boldsymbol{u}_{i}^{\text{Q}}\}_{i=1}^{n}$ from $C$, the detailed procedure is outlined in Algorithm \ref{alg3}.
Next, we describe the procedure for generating quasi-random samples of the copula $C$ using the generator network $\hat{G}$. This process consists of two stages: first, acquiring randomized QMC  points in the unit hypercube $U[0,1]^k$ using  space-filling designs (e.g. uniform designs or LHDs), and subsequently transforming these points into samples of $C$ through the composition $G_{\hat{\theta}} \circ F_{\boldsymbol{z}}^{-1}$. To produce $n$ quasi-random samples $\{\boldsymbol{u}_{i}^{\text{Q}}\}_{i=1}^{n}$ from $C$, please refer to Algorithm \ref{alg3} for the detailed process.

\begin{algorithm}[!htbp]
\caption{Training GANs. }
\label{alg2}
\begin{algorithmic}[1]
%\begin{spacing}{1.1}
\REQUIRE  a) The pseudo-observed samples $\{\boldsymbol{u}_i\}_{i=1}^{N}$ of $C$;  b) the batch size $ n_{0}$, the initial $\boldsymbol{\theta}$ and $\boldsymbol{\phi}$, where each component of them is a random number drawn from $N(0,I)$, source distribution $\nu$.
\ENSURE  Generator $\hat{G}$ and discriminator $\hat{D}$.
\WHILE  {$\boldsymbol{\theta}$ and $\boldsymbol{\phi}$ have not converged}
\FOR  {$k=1, 2, \ldots $}
\STATE  Sample minibatch of $ n_{0}$ noise samples $\{\boldsymbol{z}_{1},\ldots, \boldsymbol{z}_{n_{0}}\}$   from source distribution $\nu$.
\STATE Sample minibatch of $n_{0}$ examples $\left\{\boldsymbol{u}_{1}, \ldots, \boldsymbol{u}_{n_{0}}\right\}$ from $\{\boldsymbol{u}_i\}_{i=1}^{N}$.
\STATE  Update the discriminator by ascending its stochastic gradient:
\hspace{-5cm}
\begin{equation*}
\setlength\belowdisplayskip{0.5pt}
  \begin{aligned}
D_{\boldsymbol{\phi}}\leftarrow\nabla_{\boldsymbol{\phi}} \left[\frac{1}{n_{0}} \sum_{i=1}^{n_{0}}\log D_{\boldsymbol{\phi}}\left(\boldsymbol{u}_{i}\right)+ \frac{1}{n_{0}} \sum_{i=1}^{n_{0}}\log\Big(1-D_{\boldsymbol{\phi}}\left(G_{\boldsymbol{\theta}}\left(\boldsymbol{z}_{i}\right)\right)\Big)\right].
\end{aligned}
\end{equation*}
\STATE  Take a gradient step to update $\boldsymbol{\phi}$ with the {\rm RMSProp} optimizer by popularized \cite{Tieleman2012}.
%\STATE $\boldsymbol{\phi} \leftarrow \boldsymbol{\phi} +\alpha\cdot  {\rm RMSProp} (\boldsymbol{\phi}, f_{\boldsymbol{\phi}})$
%\STATE $\boldsymbol{\phi} \leftarrow  {\rm clip} (\boldsymbol{\phi}, -c,c)$
\STATE Sample minibatch of $ n_{0}$ noise samples $\{\boldsymbol{z}_{1}^{\prime},\ldots, \boldsymbol{z}_{n_{0}}^{\prime}\}$   from source distribution $\nu$.
\STATE Update  the generator by descending its stochastic gradient:
%\STATE Obtain a OA-LHD point set $\{\mathbf{v}_1,\cdots,\mathbf{v}_{n_{0}}\}$ from $[0,1]^p$;
%\STATE Compute $\left\{\boldsymbol{z}^{(1)}, \ldots, \boldsymbol{z}^{(n_{0})}\right\}$ by $\boldsymbol{z}^{(i)}=F^{-1}_{\boldsymbol{z}}(\mathbf{v}_i) $, where $F_{\boldsymbol{z}}$ is the distribution of the noise prior $F_{\boldsymbol{Z}}$.
%\STATE Update the generator by descending its stochastic gradient:
%\hspace{-5cm}
\begin{equation*}
\setlength\belowdisplayskip{0.3pt}
  \begin{aligned}
G_{\boldsymbol{\theta}} \leftarrow \nabla_{\boldsymbol{\theta}} \frac{1}{n_{0}} \sum_{i=1}^{n_{0}} \log \Big ( 1-D_{\boldsymbol{\phi}}\left(G_{\boldsymbol{\theta}}\left(\boldsymbol{z}_{i}\right)\right)\Big) .
\end{aligned}
\end{equation*}
\hspace{-6cm}
%\STATE $\boldsymbol{\theta}\leftarrow \boldsymbol{\theta}-\alpha \cdot {\rm RMSProp} (\boldsymbol{\theta}, G_{\boldsymbol{\theta}})$
\STATE  Take a gradient step to update $\boldsymbol{\theta}$ with the {\rm RMSProp} optimizer.
%\end{spacing}
\ENDFOR
\ENDWHILE\\
Set $\hat{G}=G_{\boldsymbol{\theta}}$ and $\hat{D}=G_{\boldsymbol{\phi}}$.
\end{algorithmic}
\end{algorithm}

\begin{algorithm}
\caption{Quasi-Random Sampling for Copulas Using GANs and Space-Filling Designs. }
\label{alg3}
\begin{algorithmic}[1]
%\begin{spacing}{1.1}
\REQUIRE  a)  the amount of generated quasi-random samples $n$;  b)  the trained generator $\Hat{G}$.
\ENSURE   $\boldsymbol{u}_{1}^{\text{Q}}, \ldots, \boldsymbol{u}_{n}^{\text{Q}}$.
%\STATE Obtain the  pseudo-observations samples $U_i=(R_{i1},\ldots, R_{id})$ of the potential $C$ via (\ref{psu}).
\STATE   Obtain the quasi-random samples  $P_{n}=\left\{\boldsymbol{v}_1, \ldots, \boldsymbol{v}_{n}\right\}$ on $U[0,1]^k$ via a space-filling designs  (e.g.,  a uniform design or an  LHD).
\STATE    Obtain the quasi-random samples $\{\boldsymbol{z}_i\}_{i=1}^{n}$ of $\nu$ by $\boldsymbol{z}_i=F_{\boldsymbol{z}}^{-1}\left(\boldsymbol{v}_i\right)$.
\STATE   Compute $\boldsymbol{u}_{i}^{\text{Q}}=\Hat{G}\left(\boldsymbol{z}_i\right), i=1, \ldots, n$.

\end{algorithmic}
\end{algorithm}

\subsection{Statistical error analysis}
This subsection aims to assess the performance of the quasi-random sampling method for estimating $\mu$ in (\ref{eq3}), i.e.
\begin{equation}\label{eq3}
 \mu=E\left[\Psi_0(\boldsymbol{X})\right]=E[\Psi(\boldsymbol{u})],
\end{equation}
where  $\boldsymbol{X} \sim F$, and  $\boldsymbol{u}$ is drawn from the copula $C$. Here the function $\Psi_0$ is  complex and computationally challenging. To approximate $\mu$,
%the MC estimator is given by:
%\begin{equation*}
%\hat{\mu}_{n}^{\mathrm{MC}}=\frac{1}{n} \sum_{i=1}^{n} \Psi\left(\boldsymbol{u}_i\right),
%\end{equation*}
%where $\boldsymbol{u}_1, \ldots, \boldsymbol{u}_{n} \stackrel{\text { i.i.d. }}{\sim} C$.
% Let $g_{\hat{\boldsymbol{\theta}}}$ denote  the generator obtaining by GANs, that is usually parameterized with a neural networks   by $\boldsymbol{\boldsymbol{\theta}}$. We train $g_{\hat{\boldsymbol{\theta}}}$ so that, given a $p$-dimensional input $\boldsymbol{Z} \sim F_{\boldsymbol{Z}}$ with independent components $Z_1, \ldots, Z_p$ from known distributions $F_{Z_1}, \ldots, F_{Z_p}$, the $g_{\hat{\boldsymbol{\theta}}}$ can generate $d$-dimensional output from the desired distribution, i.e.  $g_{\hat{\boldsymbol{\theta}}}(\boldsymbol{Z}) \sim C$. We can thus turn a space-filling design, $\left\{\tilde{\boldsymbol{v}}_1, \ldots, \tilde{\boldsymbol{v}}_n\right\}$, into a set of QRS from $C$ by letting
%\begin{equation}\label{eq2}
%\boldsymbol{u}_{i}^{\rm{Q}}=g_{\hat{\boldsymbol{\theta}}} \circ F_{\boldsymbol{Z}}^{-1}\left(\tilde{\boldsymbol{v}}_i\right), \quad i=1, \ldots, n,
%\end{equation}
%where $F_Z^{-1}(\boldsymbol{u})=\left(F_{Z_1}^{-1}\left(u_1\right), \ldots, F_{Z_p}^{-1}\left(u_p\right)\right)$.
 the associated  QMC estimator of (\ref{eq3}) is denoted as follows:
\begin{equation}\label{eq5}
 \hat{\mu}_{n}^{\mathrm{Q}}=\frac{1}{n} \sum_{i=1}^{n} \Psi\left(\boldsymbol{u}_{i}^{\rm{Q}}\right)=\frac{1}{n} \sum_{i=1}^{n} \Psi \left(\hat{G}\circ F_{\boldsymbol{z}}^{-1} (\boldsymbol{v}_i) \right),
 %\frac{1}{n} \sum_{i=1}^n \Psi\left(\boldsymbol{Y}_i\right)=\frac{1}{n} \sum_{i=1}^n \Psi\left(g_{\hat{\boldsymbol{\theta}}}\circ F_Z^{-1}\left(\tilde{\boldsymbol{v}}_i\right)\right) .
\end{equation}
where the quasi-random samples $\{\boldsymbol{u}_{i}^{\text{Q}}\}_{i=1}^{n}$ are obtained using Algorithm \ref{alg3}.
In fact, the samples $\boldsymbol{u}_{i}^{\rm{Q}}$ are derived from the push-forward distribution $\hat{G}_{\#}\nu$, which corresponds to the distribution of $\hat{G}\circ F_{\boldsymbol{z}}^{-1}$. $\hat{G}_{\#}\nu$ can be seen  as an approximation of the target copula $C$. Hence, we can express the approximation as follows:
\begin{equation*}
\mu=E(\Psi_0(\boldsymbol{X}))= E(\Psi(\boldsymbol{u})) \approx E(\Psi(\boldsymbol{u}^{\mathrm{Q}}))\approx \hat{\mu}_{n}^{\mathrm{Q}},
\end{equation*}
where $\boldsymbol{u}^{\mathrm{Q}} \sim \hat{G}_{\#}\nu$.

%In the context of evaluating $\hat{\mu}_{n_{\text {gen}}}^{\mathrm{Q}}$, we will demonstrate that if $\hat{G}$ is well-trained, the estimation error of (\ref{eq3}) using (\ref{eq5}) can be controlled by $n$ and $n$. Furthermore, we will establish the convergence rates of the bias and variance for $\hat{\mu}_{n_{\text {gen}}}^{\mathrm{Q}}$. To begin, let us introduce several notations.

In the process of assessing $\hat{\mu}_{n}^{\rm{Q}}$, we will show that when $\hat{G}$ is adequately trained, the estimation error of (\ref{eq3}) using (\ref{eq5}) can be controlled by both the sample size $N$ and the generated sample size $n$. Additionally, we will establish the convergence rates for the bias and variance of $\hat{\mu}_{n}^{\mathrm{Q}}$. For the convenience of  further discussion, let us introduce several notations.

%Define $p_{\hat{G}}$ as the density of $\hat{G}(\boldsymbol{z})$, $p_C$  as the density of the copula $C$, and  the total variation norm
%$$
%\left\|p_{\hat{G}}-p_C\right\|_{L_1}=\int_{\mathcal{X}}\left|p_{\hat{G}}(\boldsymbol{u})-p_C(\boldsymbol{u})\right| d \boldsymbol{u}.
%$$
%Now we present several theoretical properties of $\hat{\mu}_{n_{\mathrm{gen}}}^{\mathrm{GAN-QRS}}$.
%Before this, we first evaluate the learning ability of GANs, in other words, we first quantify the errors bound  of $E\left\|p_{\hat{G}}-p_C\right\|_{L_1}^2$, and then present the bias of $\mu$ and $E(\Psi(\boldsymbol{u}^{\mathrm{Q}}))$.
%%%%%%%%%%%%%%%%%%%%%
Denote the density function of the push-forward measure $\hat{G}_{\#}\nu$ as $p_{\hat{G}_{\#}\nu}$, the density function of the copula $C$ as $p_C$, and the total variation norm as follows:
$$
\left\|p_{\hat{G}_{\#}\nu}-p_C\right\|_{L_1} = \int_{\mathcal{X}}\left|p_{\hat{G}_{\#}\nu}(\boldsymbol{u})-p_C(\boldsymbol{u})\right| d \boldsymbol{u}.
$$
%In the following sections, we explore various theoretical properties of $\hat{\mu}_{n_{\mathrm{gen}}}^{\mathrm{GAN-QRS}}$. But before that, it is necessary to assess the learning capability of GANs. Specifically, we begin by quantifying the error bound of $E\left\|p_{\hat{G}_{\#}\nu}-p_C\right\|_{L_1}^2$, followed by an analysis of the bias in $\mu$ and $E(\Psi(\boldsymbol{u}^{\mathrm{Q}}))$.
In the subsequent sections, we present some  theoretical analysis  of $\hat{\mu}_{n}^{\text{Q}}$. However, it is essential to evaluate the learning prowess of GANs first. We commence by quantifying the total variation between  $p_{\hat{G}_{\#}\nu}$ and $p_C$, denoted as $E\left\|p_{\hat{G}_{\#}\nu}-p_C\right\|_{L_1}^2$, and then proceed to analyze the bias between $\mu$ and $E(\Psi(\boldsymbol{u}^{\mathrm{Q}}))$. This assessment relies on the following assumptions:
\begin{enumerate}
  \item  [(A.1)] The target  generator $G^*: \mathbf{R}^k \mapsto \mathbf{R}^d$  is continuous with $\| G^*\|_{\infty}\leq C_0$ for some constant $0<C_0<\infty$.
  \item  [(A.2)] For any $G \in \mathcal{G} \equiv \mathcal{NN} \left(W_1, L_1, B_1\right)$, $r_G(\boldsymbol{u})=p_C(\boldsymbol{u}) / \left(p_C(\boldsymbol{u})+p_{G_{\#}\nu}(\boldsymbol{u}) \right):  [0,1]^d \rightarrow [0,1]$ is continuous and $0<C_1 \leq r_G(\boldsymbol{u}) \leq C_2$ for some constants $0<C_1 \leq C_2<\infty$.
  \item  [(A.3)] The network parameters of $\mathcal{G}$ satisfies
$$
L_1 W_1 \rightarrow \infty \quad \text { and } \quad \frac{W_1^2 L_1^2 \log (W_1^2L_1^2) \log (B_1 N)}{N} \rightarrow 0, \text { as } N \rightarrow \infty.
$$
\item   [(A.4)] The network parameters of $\mathcal{D}$ satisfies
$$
L_2 W_2  \rightarrow \infty \quad \text { and } \quad \frac{W_2^2 L_2^2  \log (W_2^2L_2^2) \log (B_2 N)}{N} \rightarrow 0, \text { as } N \rightarrow \infty.
$$
\end{enumerate}

Conditions (A.1) and (A.2) are mild regularity conditions that are often assumed in nonparametric estimation problems, derived from foundational research on conditional sampling using GANs \citep{zhou2023deep}. Conditions (A.3) and (A.4) are motivated by the application of empirical process theory  \citep{vaart1997weak,Bartlett2019, zhou2023deep} to control the stochastic errors in the estimation of generators and discriminators. Specifically, conditions (A.3) and (A.4) concern the depths, widths and sizes of the generator and the discriminator networks. For the generators, these conditions require that the size of the network increases with the sample size, the product of the depth and the width increases with the sample size.

\begin{theorem}\label{th1}
 Given $N$ samples $\{\boldsymbol{u}_i\}_{i=1}^{N}$ for training a GAN, and assuming that Assumptions {\rm (A.1)--(A.4)} hold, along with the condition $N^{1/(2+d)}>d(\log N )^{1/d}$, we further make the following assumptions: {\rm (a)} the target copula $C$ is supported on $[0,1]^d$, and {\rm (b)} the source distribution $\nu$ is absolutely continuous on $\mathbf{R}^k$. Under these assumptions, we  have
\begin{equation*}
\begin{aligned}
E\| p_{\hat{G}_{\#}\nu}-p_{C}\|_{L_1}\leq \mathcal{O}\left(a_N(k,d)+b_N(k,d) \right),
\end{aligned}
\end{equation*}
where
\begin{equation}\label{aNbN}
\begin{aligned}
  a_N(k,d)&=\left(N^{-\frac{3}{8(k+2)}}  +  N^{-\frac{3}{8(d+2)}}  \right) \left(\log N\right)^{-1/4}, ~\text{and}\\
  b_N(k,d)&=N^{-\frac{(1+2k)}{4k(k+2)}}(\log N)^{(2+k)/2k} + N^{-\frac{(1+2d)}{4d(d+2)}}(\log N)^{(2+d)/2d}.
  \end{aligned}
\end{equation}
In addition, if $\| \Psi\|_{\infty}< \infty$, we can obtain
\begin{equation*}
\left|\mu - E(\Psi(\boldsymbol{u}^{\mathrm{Q}})) \right|\leq \mathcal{O}\left(a_N(k,d)+b_N(k,d) \right).
\end{equation*}
\end{theorem}
 Theorem \ref{th1}  demonstrates that, provided that the generator and discriminator networks are
suitably designed, $\hat{G}$ generated by GAN produces a good approximation distribution $\hat{G}_{\#}\nu$ of the target copula $C$.  Furthermore,  if $\| \Psi\|_{\infty}< \infty$, the bias of $\mu$ and $ E(\Psi(\boldsymbol{u}^{\mathrm{Q}}))$ will tend to 0 as $N \rightarrow \infty$.
Next, we derive an error bound between $\mu$ and   $\hat{\mu}_{n}^{\rm{Q}}$ based on Theorem \ref{th1}.

\begin{theorem}\label{th2}
Suppose that
\begin{enumerate}
    \item  [\rm{(a)}] $\Psi(\boldsymbol{u})<\infty$ for all $\boldsymbol{u} \in [0,1]^d $ and
$$
\frac{\partial^{|\beta|_1} \Psi(\boldsymbol{u})}{\partial^{\beta_1} u_1 \ldots \partial^{\beta_d} u_d}<\infty, \quad \boldsymbol{u} \in [0,1]^d,
$$
for all $\boldsymbol{\beta}=\left(\beta_1, \ldots, \beta_d\right) \subseteq\{0, \ldots, d\}^d$ and $|\boldsymbol{\beta}|_1=\sum_{j=1}^d \beta_j \leq d ;$
\item  [\rm{(b)}] there exists an $M>0$ such that $\mathrm{D}^p F_{z_j}^{-1}$ is continuous and $\left|\mathrm{D}^p F_{z_j}^{-1}\right| \leq M$ for each $p, j=1, \ldots, k$, where $\mathrm{D}^p$ denotes the $p$-fold derivative of its argument;
\item  [\rm{(c)}] for each layer $l=0, \ldots, L$ of the  $\hat{G}$, there exists an $N_l>0$ such that $\mathrm{D}^p \sigma_l$ are continuous and  $\left|\mathrm{D}^p \sigma_l\right| \leq N_l$ for all $p=1, \ldots, k $, here, $\sigma_l$ is the activation function in the $l$-th layer.
\end{enumerate}
Then, combining the conditions in Theorem \ref{th1}, we have
\begin{enumerate}
\item  [\rm{(1)}] if $P_{n}=\left\{\boldsymbol{v}_1, \ldots, \boldsymbol{v}_{n}\right\}$ is a set of random  low-discrepancy sequence on $U[0,1]^k$ (for example, a randomized Sobol sequence or a generalized Halton sequence), then
\begin{equation*}
E\left|\hat{\mu}_{n}^{ \mathrm{Q}}-\mu\right| \leq \mathcal{O}\left(a_N(k,d)+b_N(k,d)+\frac{(\log n)^{k}}{n}\right);
\end{equation*}

\item   [\rm{(2)}] if $P_{n}=\left\{\boldsymbol{v}_1, \ldots, \boldsymbol{v}_{n}\right\}$ is a  randomized OA-based LHD on $[0,1]^k$, then
\begin{equation*}
E\left|\hat{\mu}_{n}^{\mathrm{Q}}-\mu\right|\leq \mathcal{O}\left(a_N(k,d)+b_N(k,d)\right)+o\left(n^{-1/2} \right).
\end{equation*}

%\item If $P_{n}=\left\{\boldsymbol{v}_1, \ldots, \boldsymbol{v}_{n}\right\}$ is a  randomized OA-based Latin hypercube  samples on $U[0,1]^k$, which is based on an orthogonal array $O A\left(\lambda b^t, b^k,  t\right)$ and generated by algorithm, with $b \geq k \geq 2$, and $k\leq d$, then
%\begin{equation*}
%\begin{aligned}
%E\left|\hat{\mu}_{n}^{\mathrm{Q}}-\mu\right|&=\mathcal{O}\left(a_N(k,d)+b_N(k,d)+b^{-2t/k}\right)\\
%                                                                 &=\mathcal{O}\left(a_N(k,d)+b_N(k,d)+n^{-2/k}\right).
%\end{aligned}
%\end{equation*}
\end{enumerate}
Here, $a_N(k,d)$ and $b_N(k,d)$ are the same as specified in \eqref{aNbN}  of Theorem \ref{th1}.
\end{theorem}

In Theorem \ref{th2}, conditions (a)--(c) stem from the convergence theory of the GMMN QMC estimator  in \citep{hofert2021}.
%Condition (a) requires the function $\Psi$ to demonstrate global finiteness and the existence and finiteness of partial derivatives up to a $d$ order.
%Condition (a) requires the function $\Psi$ to be ?globally finite. It also
%requires that the partial derivatives of $\Psi$ up to order $d$ be well-defined and finite.
Condition (a) requires  the function $\Psi$ to be globally finite. Additionally, it requires that the partial derivatives of $\Psi$ up to order $d$ are well-defined and finite.
This condition serves as a common regularity constraint in functional space analysis, providing a foundation for subsequent theoretical derivations. The source distribution ($F_{\boldsymbol{z}}$) selected in this paper is the Gaussian distribution, thereby satisfying condition (b). Condition (c) imposes constraints on the activation function, necessitating that it exhibits smoothness. This requirement is compatible with a variety of commonly used activation functions, including, but not limited to, the Sigmoid function,  softplus function, linear activation and tangent hyperbolic  function.

Theorem \ref{th2} establishes the convergence rate of the bias of $\hat{\mu}_{n}^{\mathrm{Q}}$ to approximate $\mu$.  Subsequently, we provide the variance of  $\hat{\mu}_{n}^{\mathrm{Q}}$.
\begin{theorem}\label{th3}
Under the conditions of Theorem \ref{th2}, for two types of space-filling designs  $P_{n} \in U[0,1]^k$ to generate the quasi-random samples for copula $C\in [0,1]^d$. We have
\begin{enumerate}
\item  [\rm{(1)}] if $P_{n}=\left\{\boldsymbol{v}_1, \ldots, \boldsymbol{v}_{n}\right\}$ is a set of random  low-discrepancy sequence on $[0,1]^k$ (for example, a randomized Sobol sequence or a generalized Halton sequence), then
\begin{equation*}
\mathrm{var}(\hat{\mu}_{n}^{\mathrm{Q}})\leq\mathcal{O}\left(n^{-3}\left(\log n\right)^{k-1}\right)+\mathcal{O}\left(a_N^2(k,d)+b_N^2(k,d)\right);
\end{equation*}
\item  [\rm{(2)}] if $P_{n}=\left\{\boldsymbol{v}_1, \ldots, \boldsymbol{v}_{n}\right\}$ is a randomized   OA-based LHD on $[0,1]^k$,
% which is based on an orthogonal array $O A\left(\lambda b^t, p, b, t\right)$ and generated by algorithm, with $b \geq k \geq 2$, and $p\leq d$,
then
\begin{equation*}
\mathrm{var}(\hat{\mu}_{n}^{\mathrm{Q}})\leq\mathcal{O}\left(\frac{1}{n}\right)+\mathcal{O}\left(a_N^2(k,d)+b_N^2(k,d)\right).
\end{equation*}
\end{enumerate}
Here, $a_N(k,d), b_N(k,d)$ are the same as specified in \eqref{aNbN} in Theorem \ref{th1}.
\end{theorem}

%\begin{remark}
%By \cite{owen1994}, the variance of $\hat{\mu}_{n_{\mathrm{gen}}}^{\mathrm{MC}}$  is
%$Var(\hat{\mu}_{n_{\mathrm{gen}}}^{\mathrm{GAN-MC}})=n^{-1} \sum_{|u|>0} \sigma_u^2$, which is larger than the result in (K.1) in Theorem 3 for random quasi-random samples with $p\leq 10$, and is larger than the result in (K.2) in Theorem 3.
%\end{remark}
 Theorems \ref{th1}--\ref{th3} demonstrate that the proposed method  attains excellent learning of the target $C$ and subsequently, under suitable conditions, provides an accurate estimation for $\mu$.

 \section{Simulation Studies}
%In this section, we evaluate the effectiveness of our method  with traditional methods such as CDM and advanced GMMN  method. Both GANs and GMMN are trained on pseudo-random samples $\boldsymbol{u}_1, \ldots,\boldsymbol{u}_{n}$ from the copula $C$.  For GMMN, the detailed training process and the algorithm for generating quasi-random samples  can be referred to \cite{hofert2021}, and we also use the same architecture and choice of hyperparameters as described in \cite{hofert2021}. For GANs,  the generator is structured with a single hidden layer ($L_1=2$) having a width of 64 ($W_1=64$).
%% employing LeakyReLU as the activation function for the hidden layer and Sigmoid for the output layer.
% The discriminator is composed of two hidden layers ($L_2=3$) with widths of 256 ($W_2=256$), respectively.
%  %utilizing LeakyReLU activation for the hidden layers and Sigmoid activation for the output layer.
%
   In this section, we evaluate the effectiveness of the proposed method by comparing it with traditional methods such as CDM and the advanced GMMN method. Both GAN and GMMN are trained on pseudo-random samples $\boldsymbol{u}_1, \ldots, \boldsymbol{u}_N$ obtained via (\ref{psu}). For GMMN, we follow \cite{hofert2021} for training and sample generation, using the same architecture and hyperparameters detailed there. For GAN, the generator consists of a single hidden layer (with 64 units, $L_1=2$, $W_1=64$), while the discriminator comprises two hidden layers with 256 units ($L_2=3$, $W_2=256$).

 We employ Algorithm \ref{alg2} for GAN training and Algorithm \ref{alg3} to generate quasi-random samples $\boldsymbol{u}_{1}^{\text{Q}}, \ldots, \boldsymbol{u}_{n}^{\text{Q}}$. To employ the CDM method, we consider three copula types: Clayton, Gumbel, and the bivariate Marshall--Olkin copula.
\begin{itemize}
    \item [(a)] Clayton copula: $C(\boldsymbol{u})=\psi_C\left(\psi_{C}^{-1}\left(u_1\right)+\cdots+\psi_{C}^{-1}\left(u_d\right)\right), \quad \boldsymbol{u} \in[0,1]^d$, $\psi_C(t)=(1+t)^{-1 / \theta}($ for $\theta>0)$.
    \item [(b)] Gumbel copula: $C(\boldsymbol{u})=\psi_{G}\left(\psi_{G}^{-1}\left(u_1\right)+\cdots+\psi_{G}^{-1}\left(u_d\right)\right), \quad \boldsymbol{u} \in[0,1]^d$, $\psi_G(t)=\exp(-t^{-1 / \theta})($ for $\theta\ge 1)$.
    \item [(c)] Bivariate Marshall--Olkin copula:
    $$
C\left(u_1, u_2\right)=\min \left\{u_1^{1-\alpha_1} u_2, u_1 u_2^{1-\alpha_2}\right\}, \quad u_1, u_2 \in[0,1],
$$
where $\alpha_1, \alpha_2 \in[0,1]$.
\end{itemize}
Here, for the Clayton copula and Gumbel copula, the single parameter $\theta$ will be chosen such that Kendall's tau, denoted by $\tau$, is equal to 0.25. For the bivariate Marshall--Olkin copula, we choose $\alpha_1=0.75$ and $\alpha_2=0.60$.

\subsection{ Visual Assessments of Quasi-Random Samples Generated by GANs and Space-Filling Designs}

%Figure \ref{fig1} displays quasi-random samples  from a bivariate Marshall–Olkin copula, a three-dimension Clayton copula,  and a three-dimensional Gumbel copula, respectively. The left column of Figure \ref{fig1} shows the quasi-random samples obtaining by the method of CDM, and the right column shows the quasi-random samples  using GANs. The quasi-random samples on $U[0,1] ^ k$ used in these two methods are the randomized sobol' sequences. The similarity between the right column and  the left column indicates that the copulas $C$ are learned sufficiently well by their corresponding generator by GANs.

%In this subsection, we want to simulate the ability of GANs to learn copula. We use our method (Algorithm \ref{alg3} ) for quasi-random sampling of specific copulas. The benchmark for comparison here is to use CDM for quasi random sampling of copula.
%Figure \ref{fig1} displays quasi-random samples drawn from a bivariate Marshall--Olkin copula, a three-dimensional Clayton copula, and a three-dimensional Gumbel copula. In the left column, the quasi-random samples are generated using the method CDM, while in the right column, the quasi-random samples are generated using GANs. These methods employ randomized Sobol sequences as quasi-random samples on $U[0, 1]^k$. The visual similarity between the left and right columns implies that  GANs have learned sufficient approximations to the corresponding true copulas $C$.

\begin{figure}[ht]
\centering

% 第一行：CDM
\begin{subfigure}{\linewidth}
\centering
\rotatebox{90}{\scriptsize{~~~~~~~~~~~~~ CDM}}
\begin{minipage}[t]{0.3\linewidth}
\centering
\includegraphics[width=1.0\linewidth]{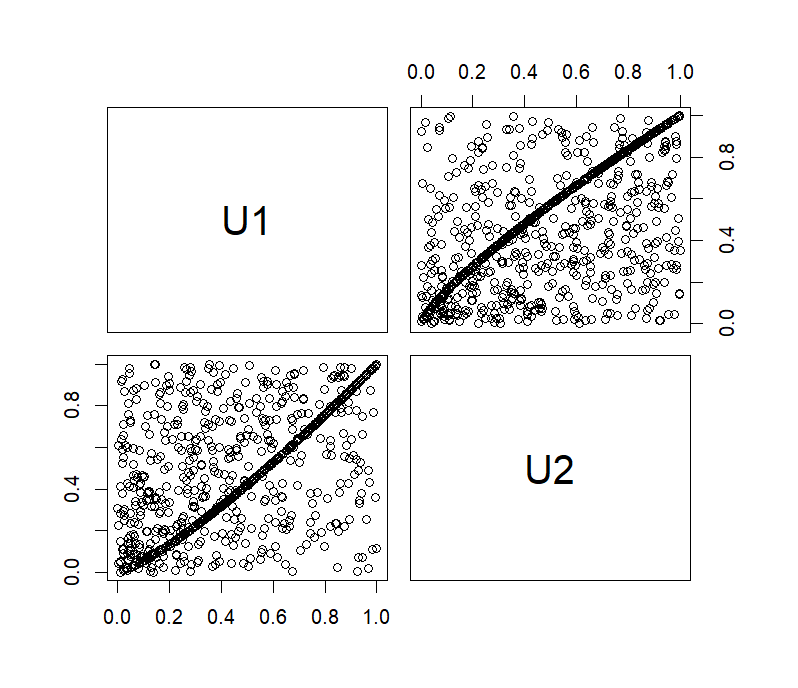}
\end{minipage}
\begin{minipage}[t]{0.3\linewidth}
\centering
\includegraphics[width=1.0\linewidth]{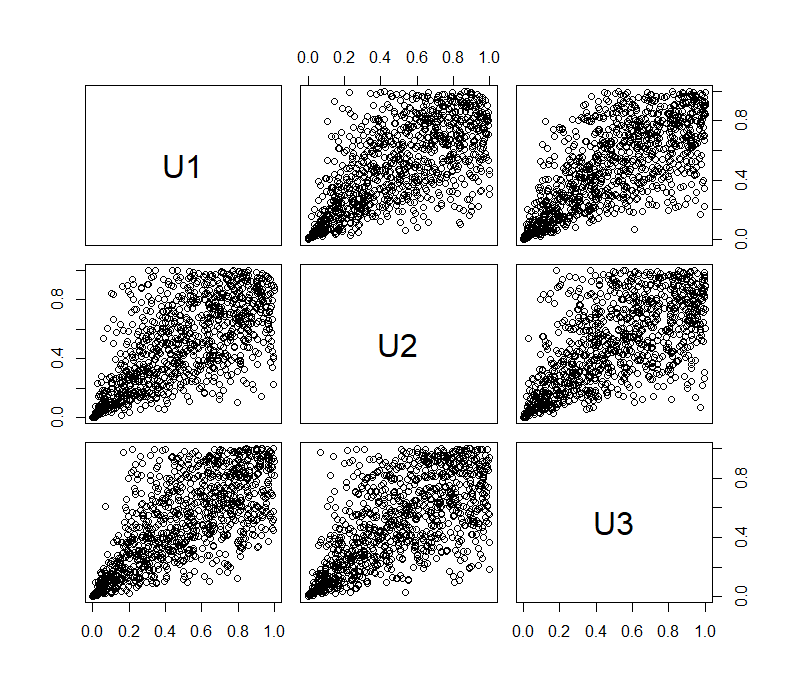}
\end{minipage}
\begin{minipage}[t]{0.3\linewidth}
\centering
\includegraphics[width=1.0\linewidth]{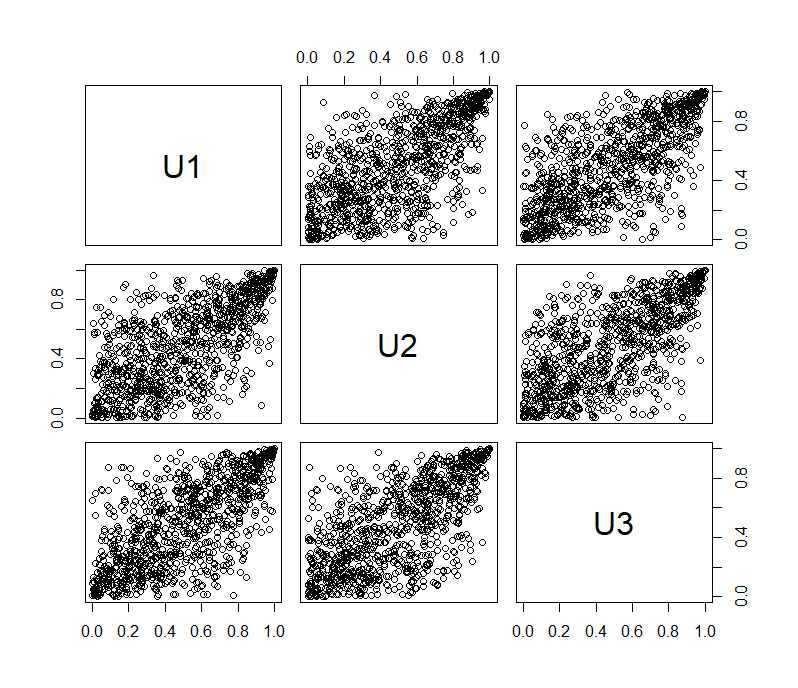}
\end{minipage}
\end{subfigure}

\vspace{0.1cm} % 两行之间的间距

% 第二行：GAN
\begin{subfigure}{\linewidth}
\centering
\rotatebox{90}{\scriptsize{~~~~~~~~~~~~ GAN}}
\begin{minipage}[t]{0.3\linewidth}
\centering
\includegraphics[width=1.0\linewidth]{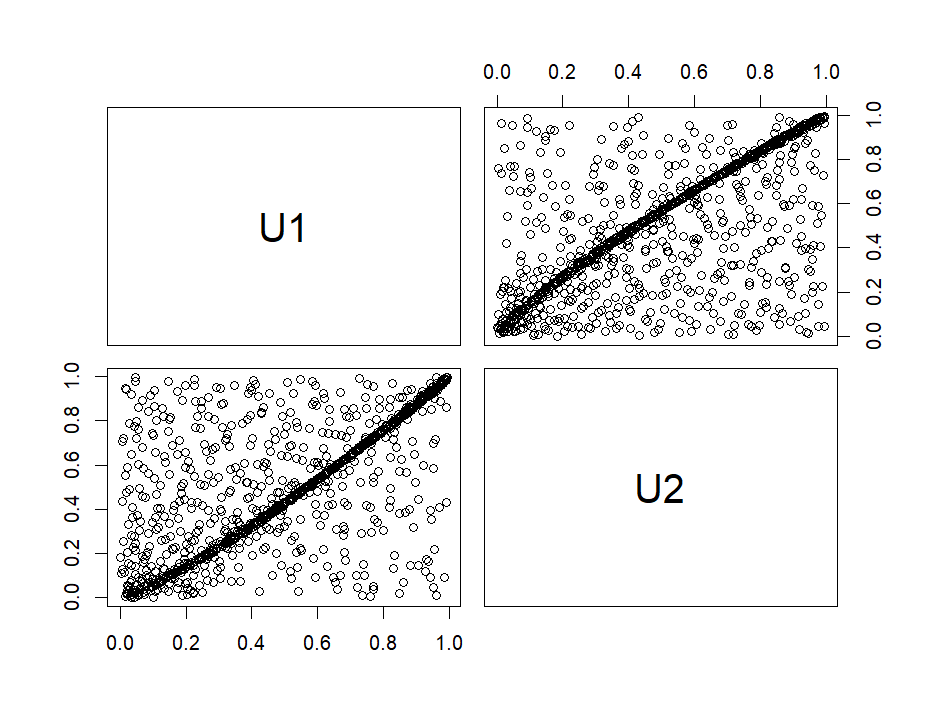}
\vspace{3pt}
\centerline{Marshall--Olkin copula}
\end{minipage}
\begin{minipage}[t]{0.3\linewidth}
\centering
\includegraphics[width=1.0\linewidth]{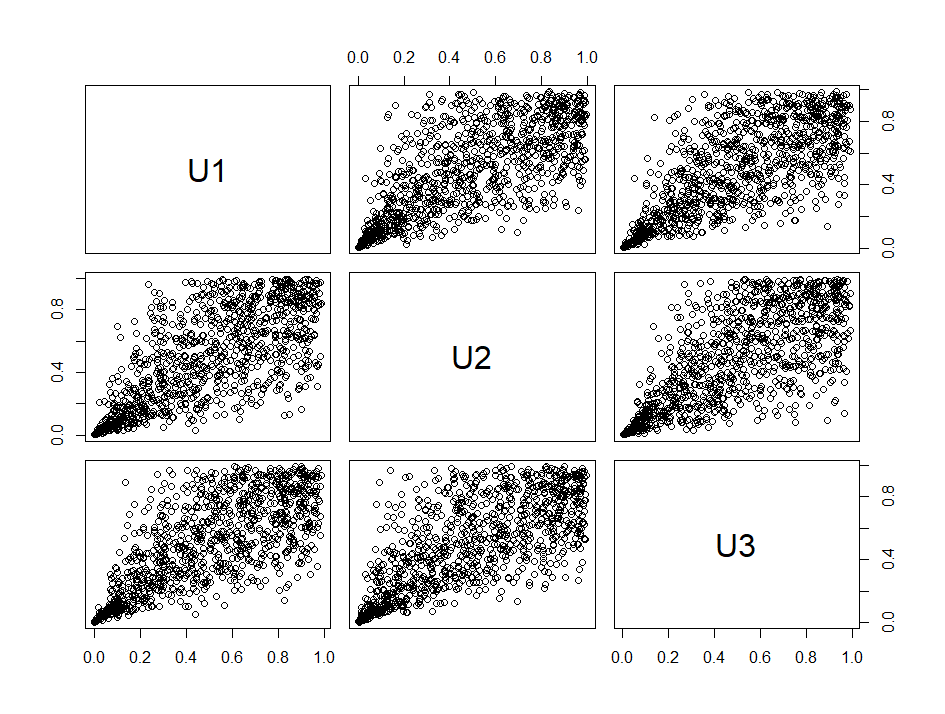}
\vspace{3pt}
\centerline{Clayton copula}
\end{minipage}
\begin{minipage}[t]{0.3\linewidth}
\centering
\includegraphics[width=1.0\linewidth]{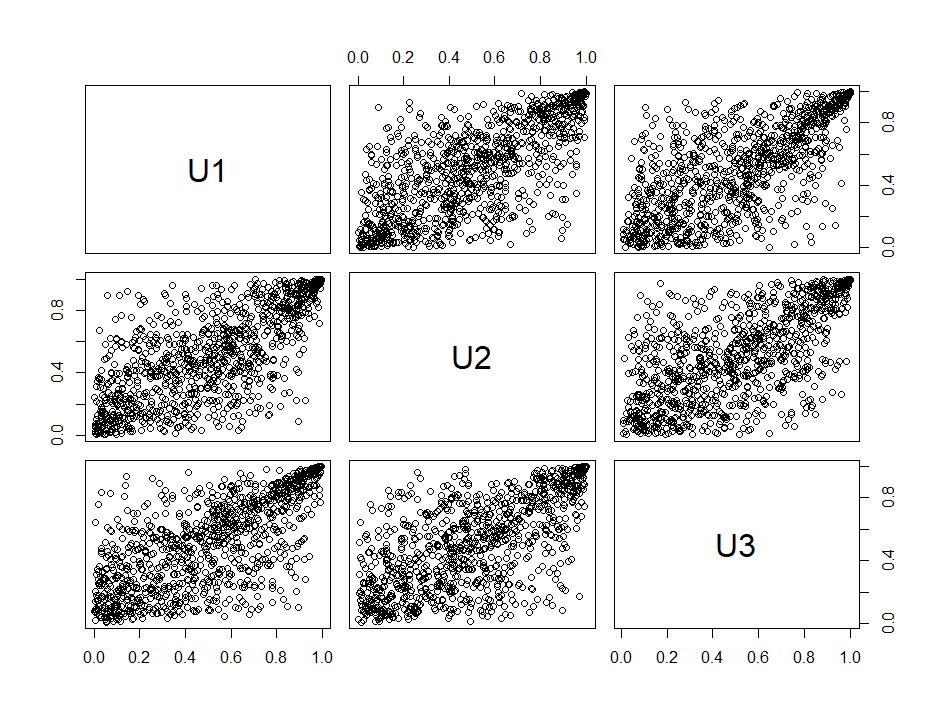}
\vspace{3pt}
\centerline{Gumbel copula}
\end{minipage}
\end{subfigure}

\caption{Quasi-random samples obtained by CDM and GAN, all of size $n=1000$, from a bivariate Marshall--Olkin copula (left), a three-dimension Clayton copula (middle) and a three-dimension Gumbel copula (right).}
\label{fig1}
\end{figure}

In this subsection, we aim to assess the capacity of GANs to learn copula models with the  method of CDM serving as the benchmark. Figure \ref{fig1} illustrates quasi-random samples derived from a bivariate Marshall--Olkin copula, a three-dimensional Clayton copula, and a three-dimensional Gumbel copula. In the top row, samples are generated using CDM's quasi-random technique, while the bottom row employs GAN-generated samples, both utilizing randomized Sobol sequences on  $U[0, 1]^k$. The visual resemblance between the columns suggests that GANs have effectively approximated the underlying true copulas.

\subsection{Assessment of  Quasi-Random Samples Generated by GANs and Space-Filling Designs via the Cram\'er-von Mises Statistic}

After visually examining the generated samples, we formally evaluate the quality of quasi-random outputs from GANs using a goodness-of-fit test.  Specifically, we employ the Cram\'er-von Mises statistic \citep{genest2009},
$$
S_{n}=\int_{[0,1]^d} n\left(C_{n}(\boldsymbol{u})-C(\boldsymbol{u})\right)^2 \mathrm{~d} C_{n}(\boldsymbol{u}),
$$
where a lower value indicates better fit. Here, the empirical copula is defined as  follows:
\begin{equation}\label{11}
  C_{n}(\boldsymbol{u})=\frac{1}{n} \sum_{i=1}^{n} \mathds{1}\left\{u_{i 1} \leq u_1, \ldots, u_{i d} \leq u_d\right\}, \quad \boldsymbol{u} \in[0,1]^d,
\end{equation}
where $\boldsymbol{u}_1, \ldots, \boldsymbol{u}_{n}$ denote quasi-random samples of the copula $C$.  We compare samples generated by three methods for three distinct copulas: the traditional CDM serves as a reference method, while GMMN and our GAN-based approach are the other two.  According to \cite{cambou2017} and \cite{hofert2021}, when employing CDM and GMMN, they recommend utilizing randomized Sobol sequences on  $U[0,1]^k$ as input for quasi-random sampling. In the proposed method, we employ both randomized Sobol sequences and  LHDs on $U[0,1]^k$ for copula sample generation.

%The traditional CDM method can be regarded as a benchmark method, while the other two methods include advanced GMMN and our GANs based approach.  When using the methods of CDM and GMMN, \cite{cambou2017} and \cite{hofert2021} suggested using randomized Sobol sequences on $U[0,1]^k$ as inputs to generate quasi-random samples. For our method, we utilize randomized Sobol sequences and randomized LHS on $U[0,1]^k$ respectively for generating the quasi-random samples of the copulas.

\begin{figure}[htbp]
	\centering
		% 第一行第一个图像
	%\subfigure{
	    \begin{minipage}[t]{0.32\linewidth}
		    \centering
		    \includegraphics[width=1.0\linewidth]{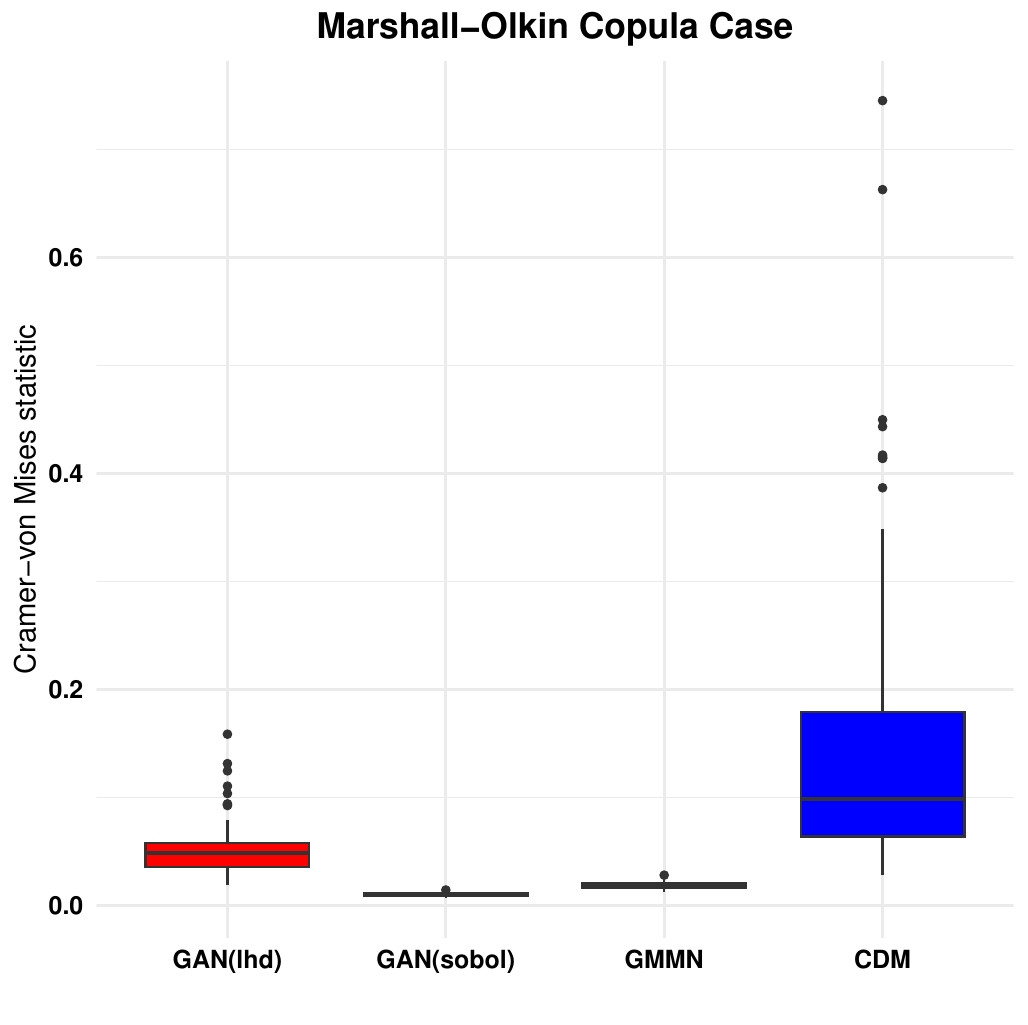}
	    \end{minipage}
	%}	
	%\subfigure{
	  %  \rotatebox{90}{\scriptsize{~~~~~~~~~~~~WD}}
	    \begin{minipage}[t]{0.32\linewidth}
		    \centering
		    \includegraphics[width=1.0\linewidth]{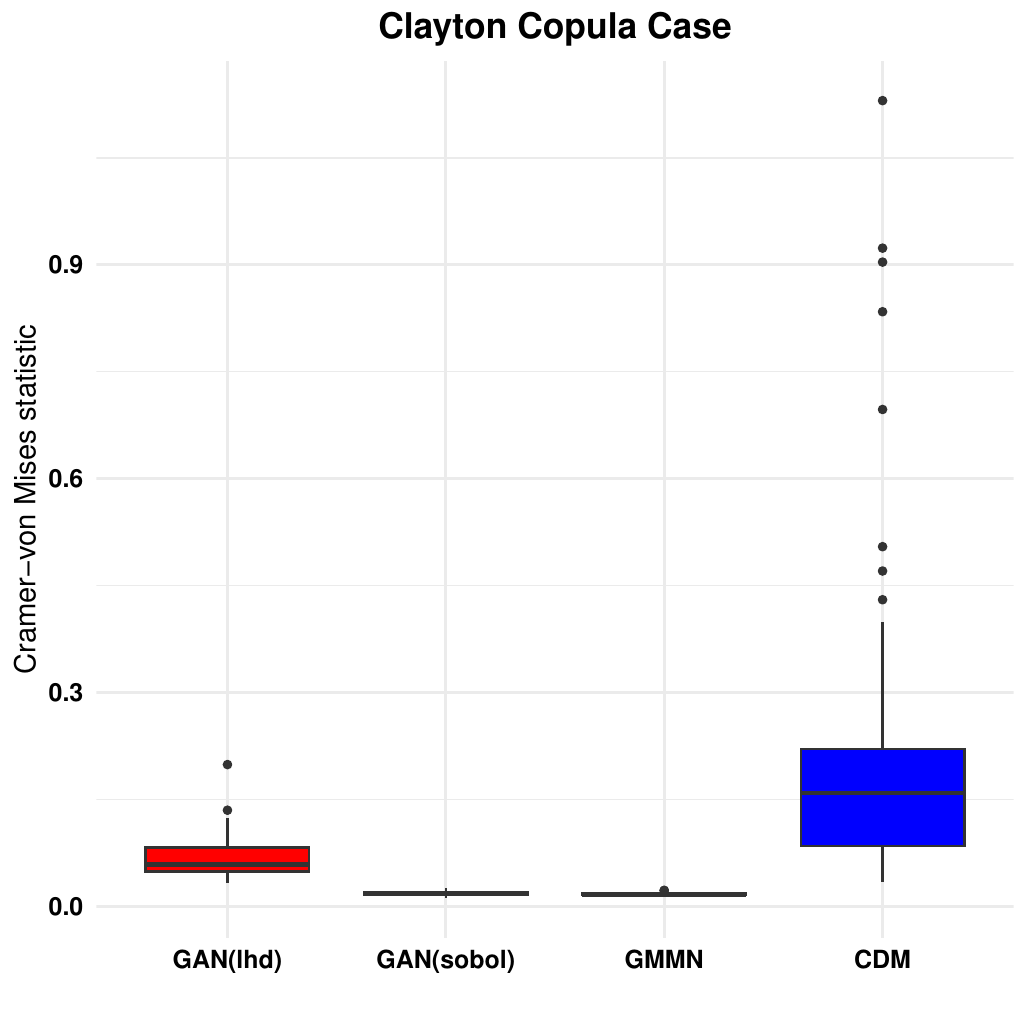}
	    \end{minipage}
	%}
  % \subfigure{
	  %  \rotatebox{90}{\scriptsize{~~~~~~~~~~~~WD}}
	    \begin{minipage}[t]{0.32\linewidth}
		    \centering
		    \includegraphics[width=1.0\linewidth]{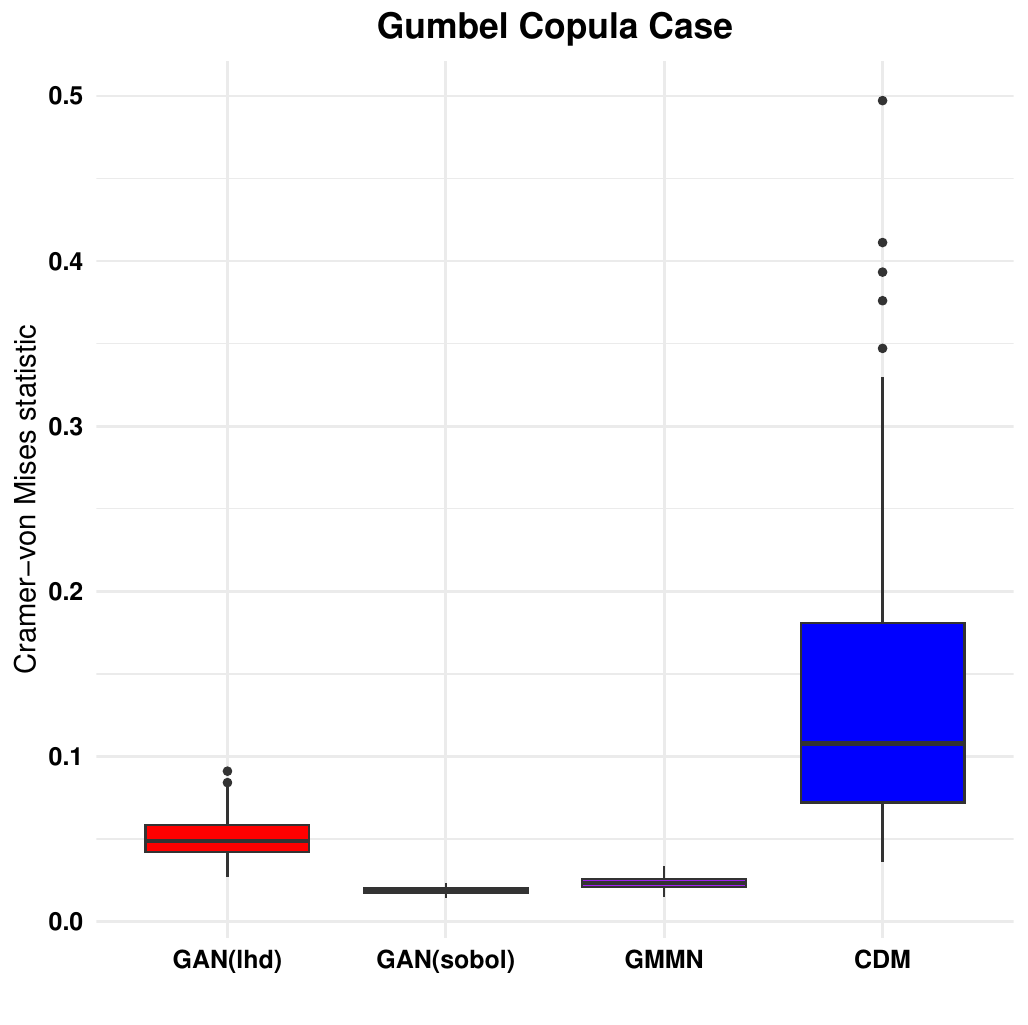}
	    \end{minipage}
	%}
  \caption{Boxplots based on $B=100$ realizations of the statistic $S_{n}$ (lower values indicate better),  constructed for three different methods: (i) the CDM, (ii) GANs with two input types, and (iii) the GMMN. All boxplots correspond to a sample size of $n=1000$. Results are displayed for a bivariate Marshall--Olkin copula (left, $d=2$), a three-dimensional Clayton copula (middle, $d=3$), and a three-dimensional Gumbel copula (right, $d=3$).}
  \label{fig2}
\end{figure}

For each copula $C$, we generate $n = 1000$ quasi-random samples and compute $B = 100$ realizations of the statistic $S_{n}$. To visualize the distribution of $S_{n}$, we utilize boxplots, as shown in Figure \ref{fig2} for a bivariate Marshall--Olkin copula (left, $d=2$), a three-dimensional Clayton copula (middle, $d=3$), and a three-dimensional Gumbel copula (right, $d=3$).
These boxplots reveal that the $S_{n}$ values derived from GAN-generated quasi-random samples are consistently lower than those from the CDM method. While the GAN method with randomized LHD inputs might be slightly less effective than GMMN, utilizing randomized Sobol sequences outperforms GMMN.

 %For each copula $C$, we generate $n=1000$ quasi-random samples, and compute $B=100$ realizations of $S_{n}$.
%  We use boxplots to depict the distribution of $S_{n}$ in each case. Figure  \ref{fig2} displays these boxplots for a bivariate Marshall–Olkin copula (left),  a three-dimension Clayton copula (middle ) and a three-dimension Gumbel copula (right).
%
%  We  can observe from the boxplots that the  $S_{n}$ values based on the GANs quasi-random samples are clearly lower than $S_{n}$ values based on the method of CDM.
%  For the GANs method, although using randomized LHS as inputs may be slightly inferior to the GMMN method, using randomized sobol' sequence as input performs better than the GMMN method.
%
  % In the case of the GANs method, utilizing a sobol' sequence as the input quasi-random samples results in even lower values of $S_{n}$ compared to the GMMN method.
Next, we conduct a numerical analysis to assess the variance reduction capability of the GAN-QRS estimator $\hat{\mu}_{n}^{\rm{Q}}$ for a function $\Psi$ associated with various copulas. Our examination involves two alternative QMC estimators, based on CDM and GMMN methods.  The choice of $\Psi$ is inspired by the practical relevance in risk management applications, particularly for modeling the dependence of portfolio risk factors, such as logarithmic returns \citep{mcneil2015}.  We now demonstrate the efficiency of $\hat{\mu}_{n}^{\rm{Q}}$ by calculating the expected shortfall (ES) of the aggregate loss, a popular risk measure in quantitative risk management.
%Next, we numerically investigate the variance reduction properties of the  estimator $\hat{\mu}_{n}^{\rm{Q}}$  for a function $\Psi$  associated with various copulas. Our comparison involves two other QMC estimators based on the CDM and GMMN methods. The function $\Psi$ under consideration is motivated by its relevance in risk management applications, particularly in modeling the dependence of portfolio risk-factor changes (e.g., logarithmic returns) as described \cite{mcneil2015}. We now demonstrate the
%efficiency of $\hat{\mu}_{n}^{\rm{Q}}$   by evaluating the expected shortfall (ES) of the aggregate loss, a widely used risk measure in quantitative risk management practice.
Specifically, if $\boldsymbol{X}=\left(X_1, \ldots, X_d\right)$  represents  a random vector of risk-factor changes with $N(0,1)$ margins, the aggregate loss is $S=\sum_{j=1}^d X_j$. The expected shortfall $\mathrm{ES}_{0.99}$ at level 0.99 of $S$ is given by
\begin{equation*}
 \begin{aligned}
\operatorname{ES}_{0.99}(S) & =\frac{1}{1-0.99} \int_{0.99}^1 F_S^{-1}(\boldsymbol{u}) \mathrm{d} \boldsymbol{u} \\
& =E\left(S \mid S>F_S^{-1}(0.99)\right)=E\left(\Psi_0(\boldsymbol{X})\right),
\end{aligned}
\end{equation*}
where $F_S^{-1}$ denotes the quantile function of $S$. Similar to the previous approach, we will utilize three copulas, i.e.  Clayton copula, Gumbel copula and Marshall--Olkin copula, to capture the interdependence among the components of $\boldsymbol{X}$.

\begin{figure}[htbp]
	\centering
		% 第一行第一个图像
	%\subfigure{
	    \begin{minipage}[t]{0.32\linewidth}
		    \centering
		    \includegraphics[width=1.0\linewidth, height=5cm]{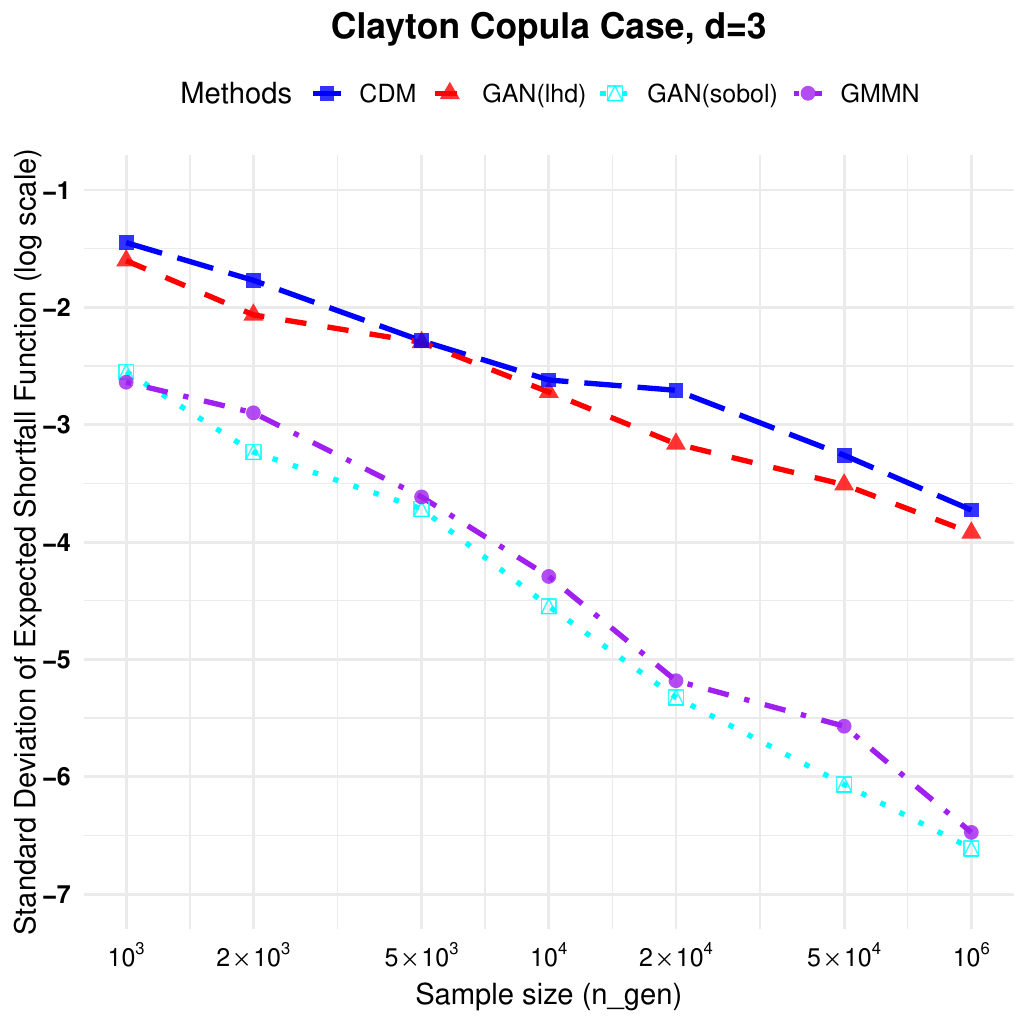}
	    \end{minipage}
	%}	
%\hfill
%	\subfigure{
	  %  \rotatebox{90}{\scriptsize{~~~~~~~~~~~~WD}}
	    \begin{minipage}[t]{0.32\linewidth}
		    \centering
		    \includegraphics[width=1.0\linewidth, height=5cm]{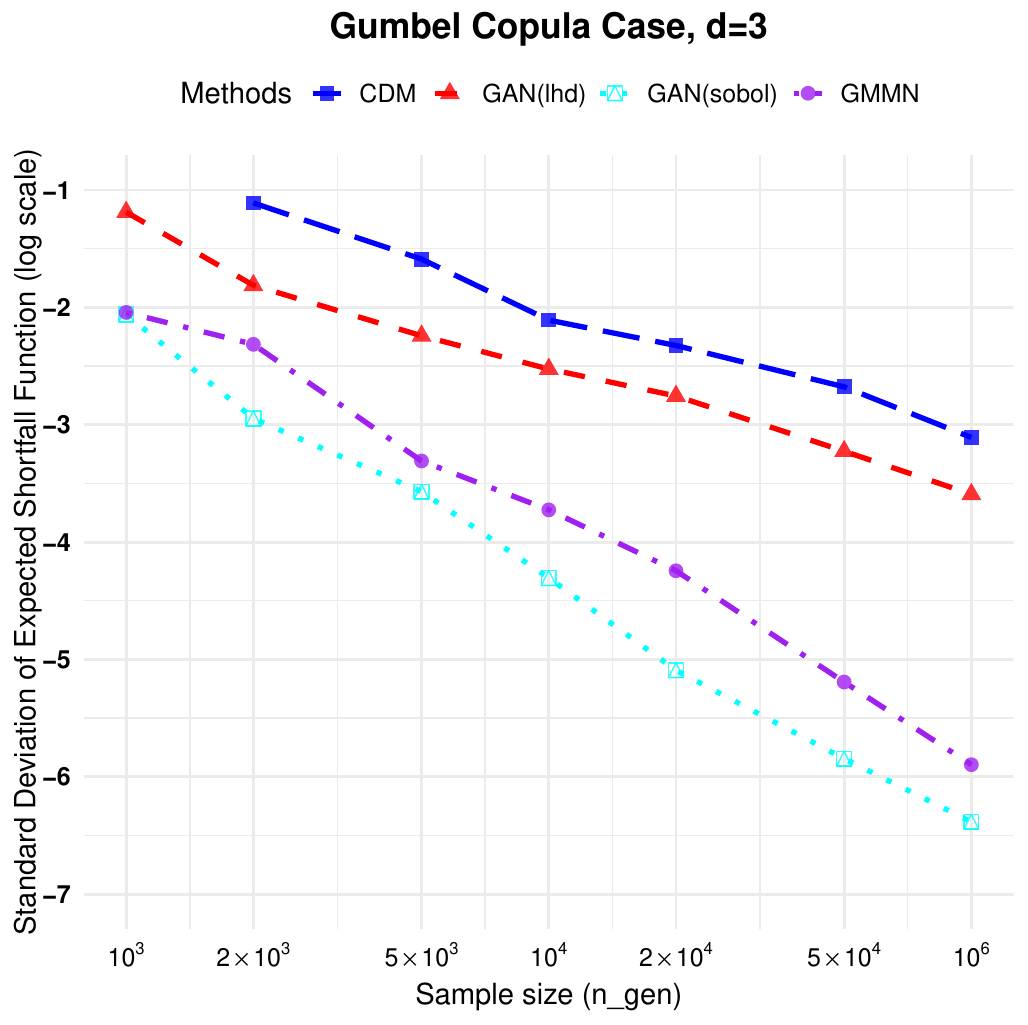}
	    \end{minipage}
%	}
%\hfill
 % \subfigure{
	  %  \rotatebox{90}{\scriptsize{~~~~~~~~~~~~WD}}
	    \begin{minipage}[t]{0.32\linewidth}
		    \centering
		    \includegraphics[width=1.0\linewidth, height=5cm]{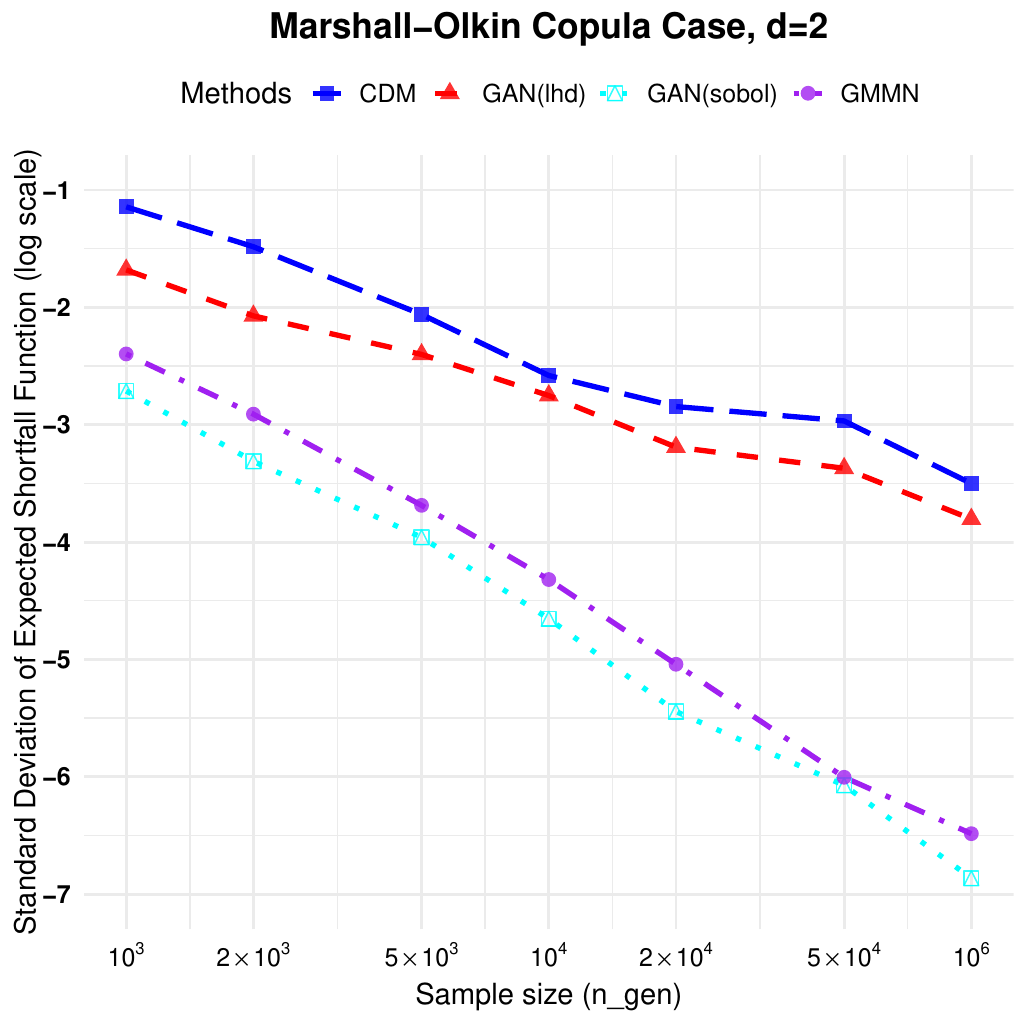}
	    \end{minipage}
	%}
  \caption{Standard deviation estimates computed using $B = 25$ replications for estimating $\operatorname{ES}_{0.99}(S)$ with CDM, GANs, and GMMN estimators,  presented for a three-dimensional Clayton copula (left), a three-dimensional Gumbel copula (middle), and a bivariate Marshall--Olkin copula (right).}
  \label{fig3}
\end{figure}

 To evaluate convergence rates, we compute standard deviation estimates using $B = 25$ randomly generated point sets $P_{n}$, with each set corresponding to a distinct value of $n$ chosen from the range $\{10^3, 2 \times 10^3, 5 \times 10^3, 10^4, 2 \times 10^4, 5 \times 10^4, 10^5\}$. This assessment aims to provide a rough estimate of convergence for all estimators.
 In this simulation study, the CDM  serves as a reference point. Figure \ref{fig3} displays the standard deviation estimates for $\mathrm{ES}_{0.99}$. A comparison is made between the estimators derived from GANs, GMMN, and the CDM. For GANs, both randomized Sobol sequences and LHDs  are employed to generate quasi-random samples. The results in Figure \ref{fig3} show that both GANs and GMMN exhibit faster convergence rates compared to traditional CDM method. It is worth noting that when GANs utilize randomized Sobol sequences, their convergence rate outperforms the GMMN.

%In this simulation study, the CDM method serves as a benchmark approach.   Figure \ref{fig3} illustrates the standard deviation estimates for  $\mathrm{ES}_{0.99}$.  A comparison is conducted between the estimators from  GANs,  GMMN, and  the CDM method.  For the GANs technique, randomized Sobol sequences and randomized LHS are employed to generate quasi-random samples. The results depicted in Figure \ref{fig3} indicate that both GANs and GMMN exhibit faster convergence rates in contrast to traditional CDM methods. Notably, it is observed that when utilizing a randomized Sobol sequence as input for GANs, its convergence rate surpasses that of GMMN.

%Here, the randomized LHS is fellow from \citep{McKay1979}.

%Figure \ref{fig3} demonstrates that both the GANs and GMMN methods exhibit faster convergence speeds compared to traditional CDM methods. Furthermore, we observed that when the input for GANs is a randomized Sobol sequence, its convergence rate surpasses that of GMMN.

\section{ Real Data Analysis}

In this section, we showcase the practical utility of the proposed method through a real-data example in finance and risk management. The focus is on modeling dependent multivariate return data, which is essential for estimating  $\mu$. GANs offer two key advantages in this context: first, they exhibit high flexibility, allowing them to capture complex dependence structures that may be inadequately represented by traditional parametric copula models  \citep{hofert2018}. Second, by generating quasi-random samples, GANs can reduce variance in estimating $\mu$. The GANs's architecture and hyperparameters are as described in Section 4.

To illustrate these benefits, we apply our method to model S$\&$P 500 constituent portfolios, with daily adjusted closing prices.
To evaluate the method's scalability to high-dimensional settings, we consider three portfolios with  $d=10$, $d=20$ and $d=200$ assets, respectively, covering the sample period from  1995-01-01 to 2015-12-31. All data are obtained from the R package `qrmdata'. 

 For the proposed method, the latent dimension $k$ plays a central role in both theoretical guarantees and practical performance, especially when $k<d$.  To clarify whether reducing $k$ yields tangible benefits, we conduct systematic experiments across all dimensionality levels: for $d=10$, we evaluate three latent dimensions: $k=3,5, 10$; for $d=20$, we test $k=5,10, 20$; for the new high-dimensional case $d=200$, we consider $k=64,128$.

To account for temporal dependencies, we adopt the copula-GARCH model \citep{jondeau2006}, using ARMA$(1,1)$-GARCH$(1,1)$ with standardized $t$-distributed innovations. We compute pseudo-observations for each portfolio, as in (\ref{psu}). These are then used to model the dependence among the log-return series.

%In order to incorporate temporal dependencies in the marginal distributions, we adopt the copula-GARCH approach \citep{jondeau2006, patton2006}. This involves modeling each individual time series of log-returns using an ARMA $(1,1)-\operatorname{GARCH}(1,1)$ model with standardized $t$-distributed innovations. Subsequently, we compute pseudo-observations, as outlined in  (\ref{psu}), for each of the two portfolios. These pseudo-observations are utilized to model the dependence among the log-return series of the respective portfolios.

 In order to generate quasi-random samples using the CDM method, we initially model each of the three portfolios with well-established parametric copulas, specifically Gumbel or uncorrelated $t$ copulas. The fitting of these copulas is carried out using the maximum pseudo-likelihood method, as described in \cite{hofert2018}. For GANs and GMMN, we first employ pseudo-observations to train the generator models. Next, we generate quasi-random samples from $U[0,1]^k$ to produce potential copulas quasi-random samples for these portfolios. Both the CDM and GMMN methods employ randomized Sobol sequences on $U[0,1]^k$ as input. In particular, we choose $k=d$, which follows recommendations in \cite{cambou2017} and \cite{hofert2021}. For GANs based method with $k=d$, we employ two types of quasi-random samples from $U[0,1]^k$: one is a randomized Sobol sequence, and the other is  a randomized OA-based LHD  with  strength  $2$. 
 %Here, we set $k=d$, where $d$ is the dimensionality of the portfolios.
 To more concisely illustrate the scenario where the input noise dimension $k < d$, the proposed GAN-based methods employ randomized OA-based LHDs as quasi-random samples on $\mathcal{U}[0,1]^k$. In particular, for the high-dimensional case $d=200$, we only present the fitting results under the Gumbel copula. This is because the values of $S_{N,n}$ obtained under the $t$-copula specification are substantially larger than those from all other competing methods.

%In order to generate quasi-random samples using the CDM method, we initially employ well-established parametric copulas to model each of the two portfolios. Specifically, we consider two types of parametric copulas: Gumbel copula and unstructured $t$ copula. The fitting of these parametric copulas is carried out using the maximum pseudo-likelihood method, as described in \citep{hofert2018}.  For the GANs and GMMN methods, we first utilize the pseudo-observations to train the corresponding generator models. Subsequently, we employ quasi-random samples from $U[0,1]^k$ to generate quasi-random samples of potential copulas for these two portfolios. The randomized Sobol sequences are used as quasi-random samples for both the CDM and GMMN methods, following the recommendations provided by  \cite{cambou2017} and \cite{hofert2021}. In the case of GANs, we introduce two types of quasi-random samples from $U[0,1]^k$: one is a randomized Sobol sequence, and the other is an orthogonal array (OA)-based Latin hypercube sampling (LHS) with an OA chosen to have a strength of 2. Here, we select $k=d$, where $d$ represents the dimensionality of the portfolios.

To assess the model fit, we  use  the empirical Cram\'er-von Mises type test statistic \citep{remillard2009} to compare the equality of two empirical copulas. This statistic is defined as
$$
S_{N,n}=\int_{[0,1]^d}\left(\sqrt{\frac{1}{n}+\frac{1}{N}}\right)^{-1}\left(C_{n}(\boldsymbol{u})-C_{N}(\boldsymbol{u})\right)^2\mathrm{~d} \boldsymbol{u},
$$
where $C_{n}(\boldsymbol{u})$ and $C_{N}(\boldsymbol{u})$ denote empirical copulas, as defined in  (\ref{11}), derived from the $n$ samples generated from the fitted dependence model and the $N$ pseudo-observations corresponding to the  training data, respectively.  The  evaluation of $S_{N, n}$ can be seen in \cite{remillard2009}.
 For each portfolio, we generate $B=20$ instances of $S_{N, n}$ using $n=1000$ quasi-random samples from the fitted model and $N=5287$ pseudo-observations. 
 
\begin{figure}[htbp]
    \centering
    
    % 图1：d=10
    \begin{subfigure}{0.63\linewidth}
        \centering
        \includegraphics[width=1.0\linewidth, height=6.5cm]{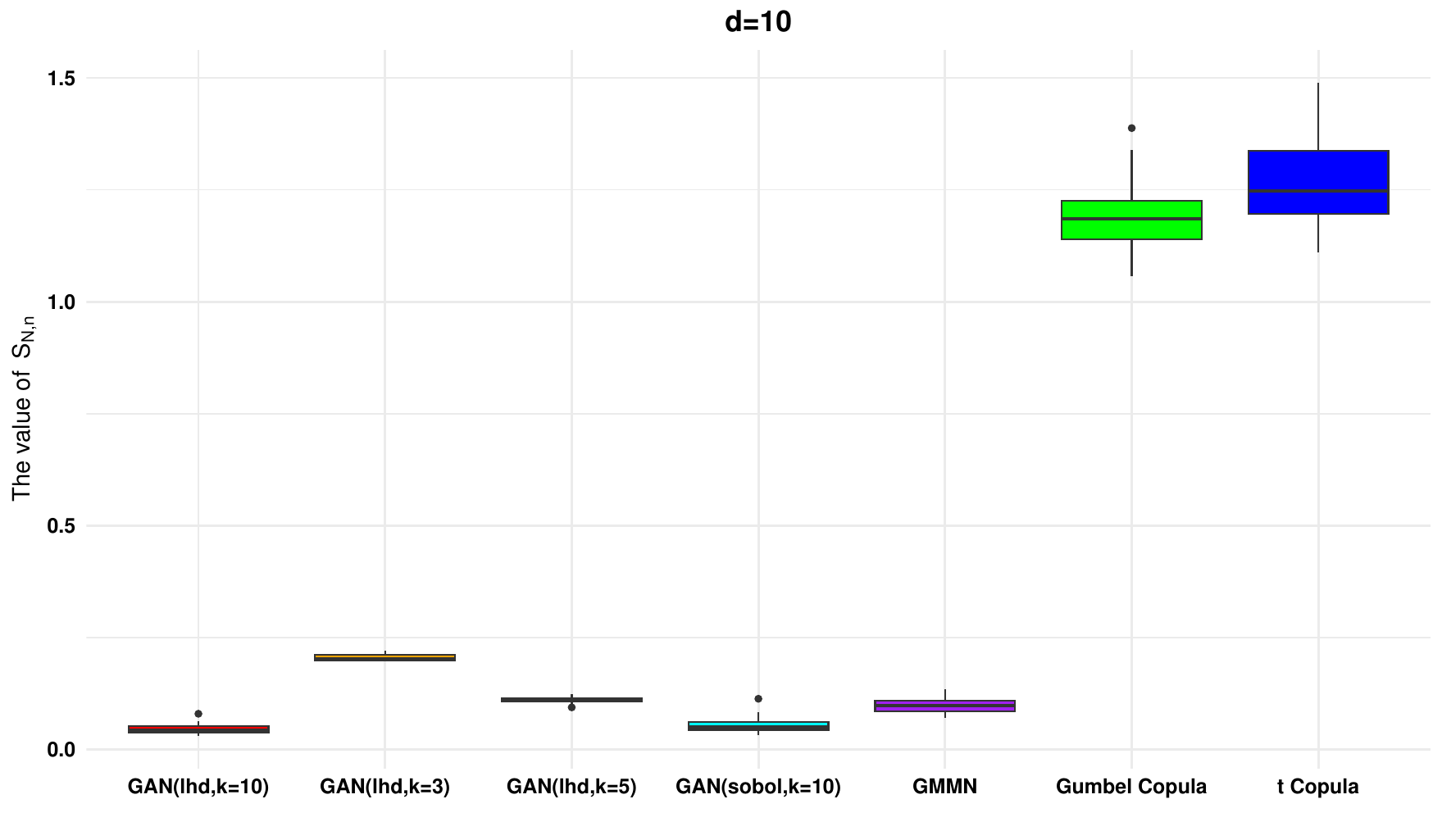}
    \end{subfigure}
    \hfill
    % 图2：d=20
    \begin{subfigure}{0.63\linewidth}
        \centering
        \includegraphics[width=1.0\linewidth, height=6.5cm]{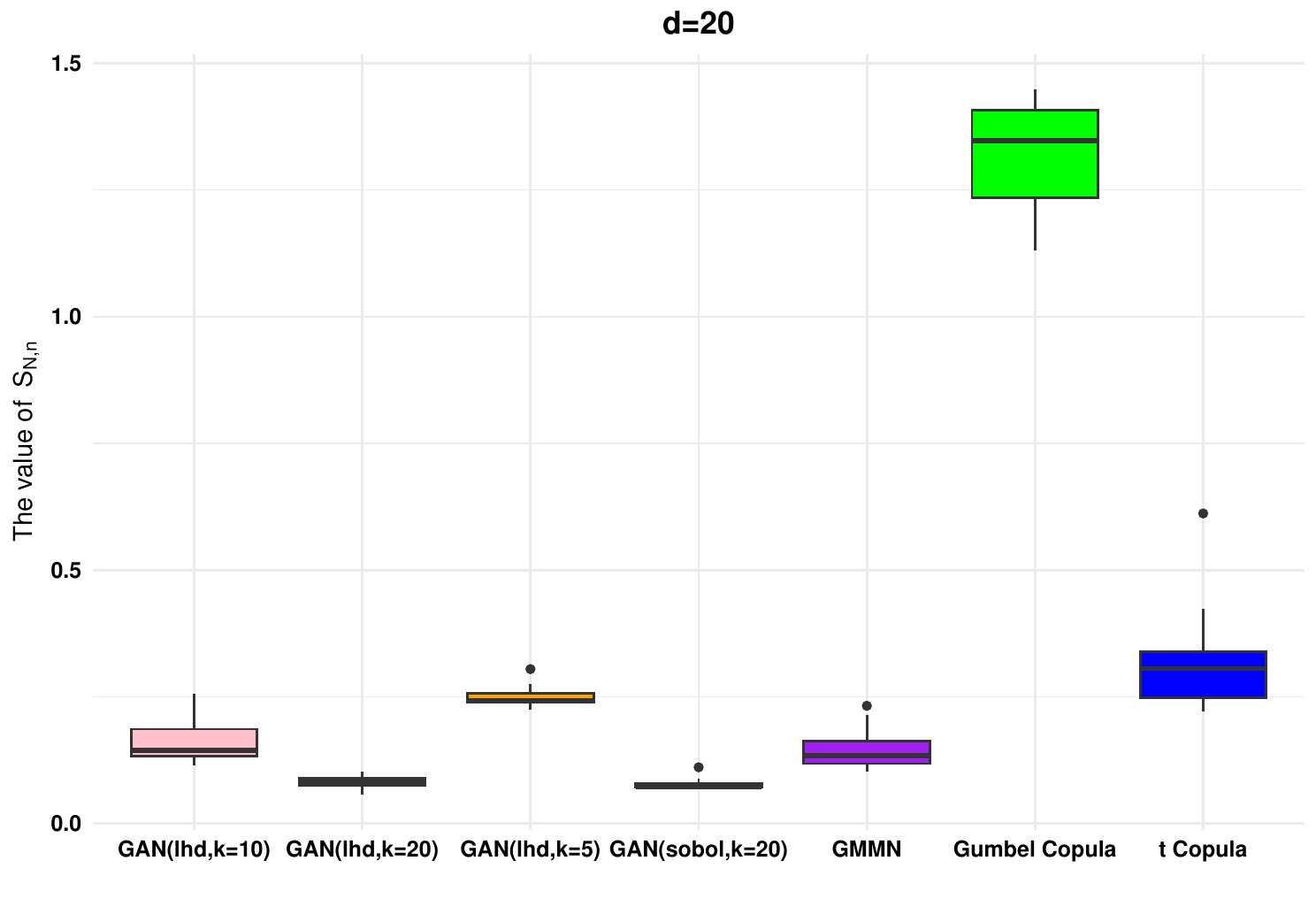}
    \end{subfigure}
    \hfill
    % 图3：d=200
    \begin{subfigure}{0.36\linewidth}
        \centering
        \includegraphics[width=1.0\linewidth, height=5.5cm]{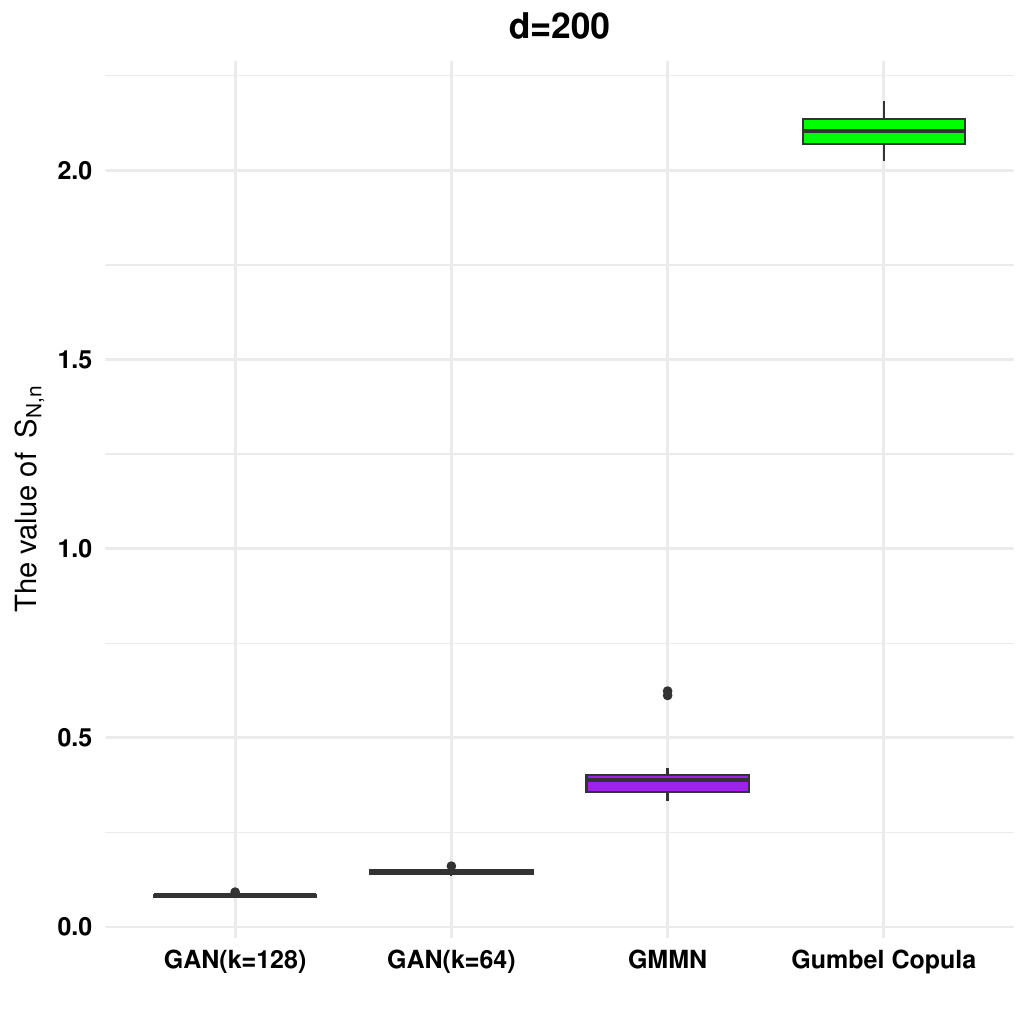}
    \end{subfigure}

    \caption{Boxplots based on $B=20$ realization of $S_{N, n}$, computed from (i) the CDM method, (ii) GANs with two types of inputs, and (iii) the GMMN method--all of the size $n=1000$ samples--for $d=10$, $d=20$, and $d=200$.}
    \label{figr1}
\end{figure}

 The boxplots in Figure \ref{figr1} show  the distribution of $S_{N, n}$ for each portfolio. 
 %%%
 It is clear that, for the case $k=d$, the distribution of $S_{N,n}$ obtained by the proposed method is more concentrated around zero compared with both the CDM method and the GMMN method. Furthermore, for scenarios where $k < d$, we observe persistent advantages of the proposed method. For example, for $d=10$, the proposed method with $k=3$ and $k=5$ still exhibits clear superiority over the CDM method. Relative to the GMMN method, while the proposed method with $k=3$ shows slightly inferior performance, it achieves comparable results at $k=5$, and demonstrates a notable advantage over the GMMN method when $k=10$.
%%%
For $d=20$ and $d=200$, analogous patterns further  confirm the scalability and robustness of the proposed method.

% Next, we examine the variance reduction effect of our GANs estimator $\hat{\mu}_{n}^{\mathrm{Q}}$ in comparison to the QMC estimator based on CDM and GMMN estimators. For the CDM and GMMN methods, we utilize randomized Sobol sequences as the quasi-random samples on $U[0,1]^k$. On the other hand, for the GANs method, we employ two types of quasi-random samples on $U[0,1]^k$, namely randomized Sobol sequences and OA-based LHS. The focus of our application is on estimating the expected shortfall $\mu=\mathrm{ES}_{0.99}(S)$ for $S=\sum_{j=1}^d X_j$ as discussed in Section 4.2. This estimation task is a common practice in risk management, as outlined in the Basel guidelines.

 Next, we investigate the variance reduction achieved by our GANs estimators, $\hat{\mu}_{n}^{\rm{Q}}$, compared to QMC estimators  using CDM and GMMN. For CDM and GMMN, we generate quasi-random samples using randomized Sobol sequences on $U[0,1]^k$. In contrast, for GANs, we employ two quasi-random sampling methods on $U[0,1]^k$: randomized Sobol sequences and randomized OA-based LHD. Our application focuses on estimating the expected shortfall, $\mu = \text{ES}_{0.99}(S)$, for the portfolio sum $S = \sum_{j=1}^d X_j$.

\begin{figure}[htbp]
    \centering
    
    % 图1：d=10
    \begin{subfigure}{0.32\linewidth}
        \centering
        \includegraphics[width=1.0\linewidth, height=6cm]{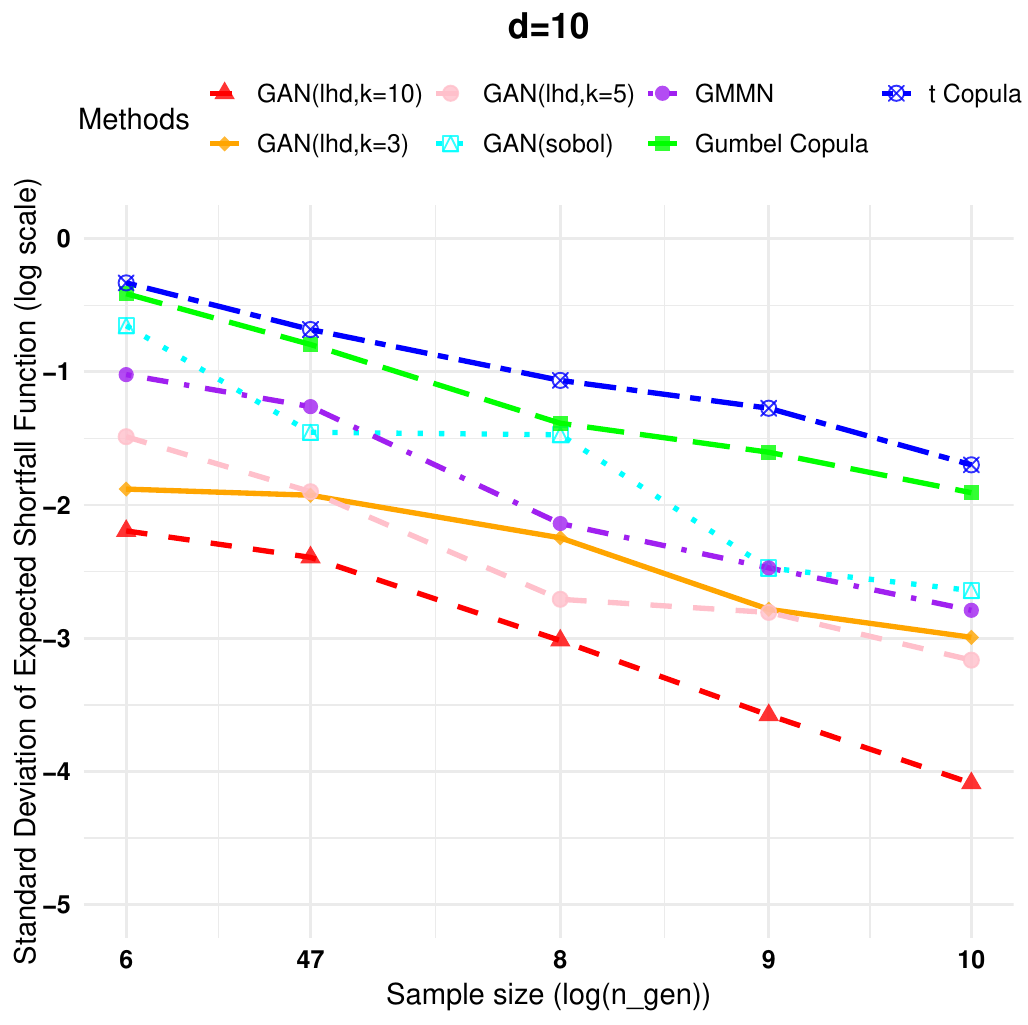}
    \end{subfigure}
    \hfill
    % 图2：d=20
    \begin{subfigure}{0.32\linewidth}
        \centering
        \includegraphics[width=1.0\linewidth, height=6cm]{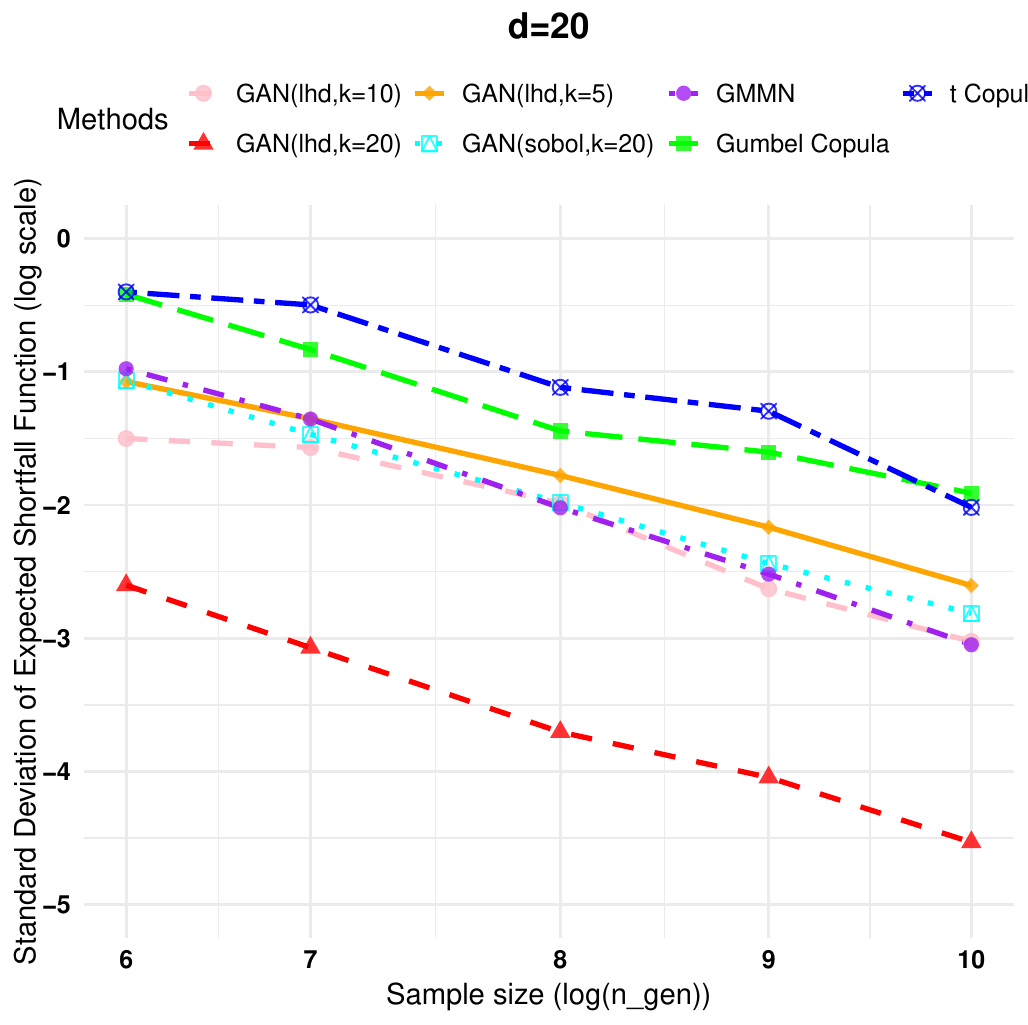}
    \end{subfigure}
    \hfill
    % 图3：d=200
    \begin{subfigure}{0.32\linewidth}
        \centering
        \includegraphics[width=1.0\linewidth, height=6cm]{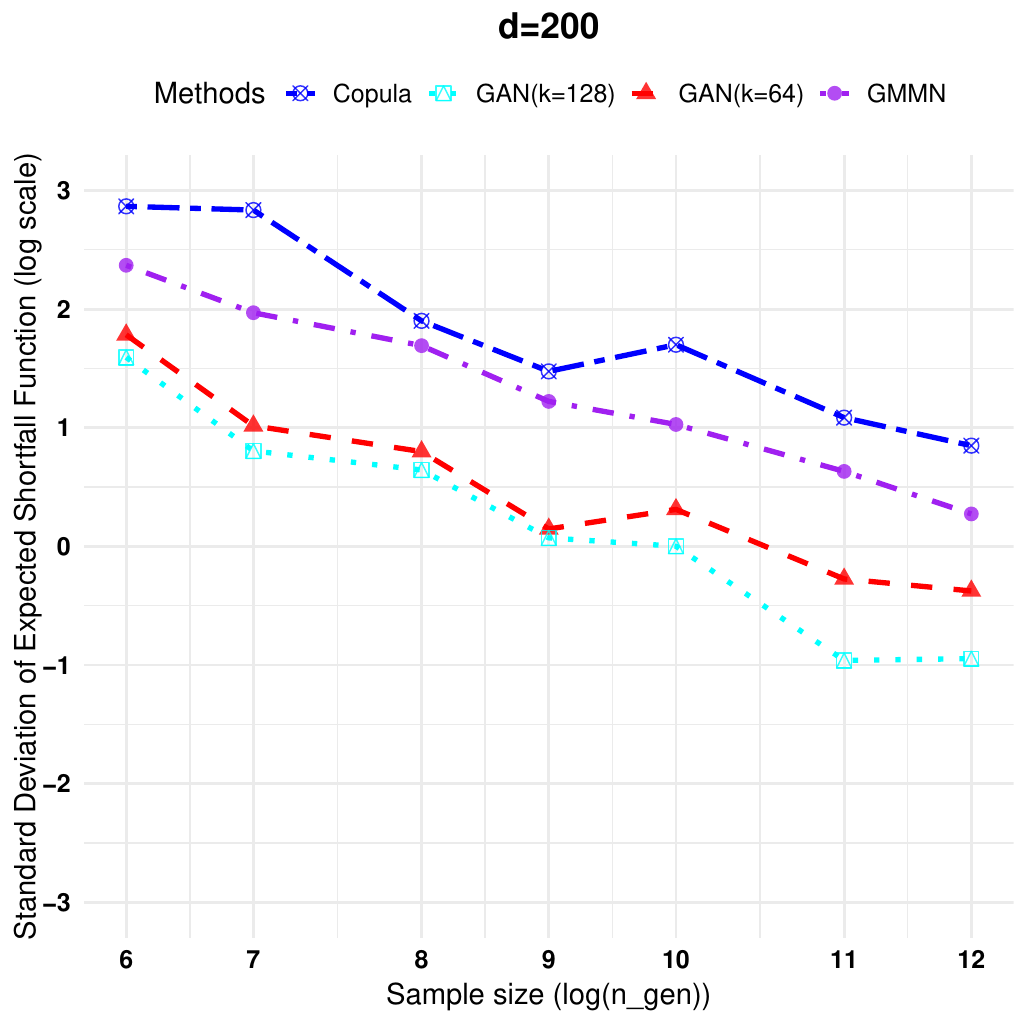}
    \end{subfigure}

    \caption{Standard deviation estimates derived from $B = 20$ replications for calculating the $\operatorname{ES}_{0.99}(S)$ estimates using CDM estimators, GANs estimators, and GMMN estimators --- for dimensions $d=10$, $d=20$ and $d=200$.}
    \label{figr2}
\end{figure}

For each portfolio with dimensions $d=10$, $d=20$ and $d=200$ respectively, we compute 20 realizations of the QMC estimators for $\mu$ using CDM, GANs, and GMMN, with the number of generated samples being denoted by $n$. The orthogonal array used to generate OA-based LHD is chosen as $OA(n, s^k, 2)$, where $n=s^2$ and $s$ is a prime number. Here, we choose $n \in \{31^2, 43^2, 67^2, 97^2, 139^2\}$. Figure \ref{figr2} illustrates the standard deviation estimates for estimating $\text{ES}_{0.99}$ for these three portfolios. 
 These plots reveal that the proposed GAN-based method consistently yields lower variance than the CDM method across all dimensionality and sample size settings.
When employing OA-based LHDs as quasi-random input noise, our method further outperforms the GMMN method. Notably, this performance advantage remains intact even for cases where $k < d$, confirming the robustness and reliability of our method across different latent dimension choices. 
%%
%\begin{figure}[htbp]
%	\centering
%	
%	% 第一行第一个图像
%	\subfigure{
%	    \begin{minipage}[t]{0.5\linewidth}
%		    \centering
%		    \includegraphics[width=1.0\linewidth, height=7cm]{d10.pdf}
%	    \end{minipage}
%	}	
%	\subfigure{
%	  %  \rotatebox{90}{\scriptsize{~~~~~~~~~~~~WD}}
%	    \begin{minipage}[t]{0.45\linewidth}
%		    \centering
%		    \includegraphics[width=1.0\linewidth, height=7cm]{d20.pdf}
%	    \end{minipage}
%	}
%\subfigure{
%	  %  \rotatebox{90}{\scriptsize{~~~~~~~~~~~~WD}}
%	    \begin{minipage}[t]{0.45\linewidth}
%		    \centering
%		    \includegraphics[width=1.0\linewidth, height=7cm]{d200.pdf}
%	    \end{minipage}
%	}
%  \caption{ Standard deviation estimates derived from $B = 20$ replications for calculating the $\operatorname{ES}_{0.99}(S)$ estimates using CDM  estimators, GANs  estimators, and GMMN  estimators - for dimensions $d=10$ (left) and $d=20$ (right).}
%  \label{figr2}
%\end{figure}

\section{Concluding  Remarks}\label{sec6}
This paper addresses the critical question of how to obtain quasi-random samples for a diverse range of copulas. Until recently, such sampling was feasible only for a limited number of copulas with specific structures. To overcome these limitations, we propose a computationally efficient quasi-random sampling framework that offers strong theoretical guarantees and synergistically integrates GANs with space-filling designs. The proposed approach first utilizes GANs to learn the optimal transformation $\phi_{C}$, which establishes a precise mapping between low-dimensional uniform distributions and high-dimensional copula structures. By subsequently employing space-filling designs to generate randomized QMC points in low-dimensional uniform spaces and mapping these points to the target copula via $\phi_{C}$, the proposed method achieves superior performance across multiple dimensions. The proposed framework demonstrates three key advantages: (a) optimal dimensionality reduction through $\phi_{C}$, which eliminates the need for complex high-dimensional space-filling designs; (b) enhanced performance compared to traditional CDM method and GMMN, particularly in high dimensions and with limited data; and (c) theoretical guarantees for low-variance MC estimators through bias and variance bounds. Empirical results and a real data analysis  validate its universality and efficiency.

Looking ahead, there still remain two main challenges for quasi-random sampling in higher-dimensional copulas. First, we must acknowledge that GANs are not the most popular generative models at present. Improving the learning of complex distributions through other generative models, such as diffusion models, is a sustainable area of research. Additionally, it is desirable to propose a novel approach for generating quasi-random samples within a hierarchical framework of conditional copulas, which are commonly encountered in image generation.

%%%%%%%%%%%%%%%%%%%%%%%%%%%%%%%%%%%%%%%%%%%%%%%%%%%%%%%%%%%%%%%%%%%%%%%%%%%%%%%%%%%%%%%%%%%%%%%%%%%%%%%%%%%%%%%%%%%%%%%%%%%%
\section*{Supplementary Materials}

The supplementary materials provides the  proofs of the statistical error analysis for this paper.
%\begin{description}
%	
%	\item[Online Appendix:] provides the  proofs of the statistical error analysis for this paper.
%	
%
%	\item[Code:] provides the Python code and related data for this paper, see \url{https://github.com/SuminWang-088/GANs-QRS-copula}.
%
%
%
%	%\item[Code:] provides the Python code and related data for this paper. %, see \url{https://github.com/SuminWang-088/GANs-QRS-copula}.
%
%
%	
%	
%\end{description}
%%%%%%%%%%%%%%%%%%%%%%%%%%%%%%%%%%%%%%%%%%%%%%%%%%%%%%%%%%%%%%%%%%%%%%%%%%%%%%%%%%%%%%%%%%%%%%%%%%%%%%%%%%%%%%%%%%%%%%%%%%%%
\section*{Acknowledgements}

This work was supported by the National Natural Science Foundation of China (Grant Nos.~12401324, 12131001 and 12371260), the  Social Science Foundation of Hebei Province (Grant No.~HB25TJ003), and the Hebei  Provincial Statistical Research  Project (Grant No.~2025HY26). The first two authors contributed equally to this work.
\par

%%%%%%%%%%%%%%%%%%%%%%%%%%%%%%%%%%%%%%%%%%%%%%%%%%%%%%%%%%%%%%%%%%%%%%%%%%%%%%%%%%%%%%%%%%

\bibhang=1.7pc
\bibsep=2pt
\fontsize{9}{14pt plus.8pt minus .6pt}\selectfont
\renewcommand\bibname{\large \bf References}
%\begin{thebibliography}{11}
\expandafter\ifx\csname
natexlab\endcsname\relax\def\natexlab#1{#1}\fi
\expandafter\ifx\csname url\endcsname\relax
  \def\url#1{\texttt{#1}}\fi
\expandafter\ifx\csname urlprefix\endcsname\relax\def\urlprefix{URL}\fi

%% use bibfile 
%  \bibliographystyle{chicago}      % Chicago style, author-year citations
%  \bibliography{bibfile}   % name your BibTeX data base
 \bibliographystyle{chicago} 
 \bibliography{bibliography.bib}

%%%  Another method
%\begin{thebibliography}{}
%
%\bibitem[Curtis(1943)]{1943}
%Curtis, M. (1943).
%{\em Documents on International Affairs, 1938}, Volume~II.
% Oxford University Press, London.
%
%\bibitem[Eubank(2004)]{Eubank}
%Eubank, K. (2004).
% {\em The origins of World War II}. 
% 3rd Edition.
%Harlan Davidson, Wheeling, Ill.
%
%\bibitem[Gellately(1988)]{1988}
%Gellately, R. (1988).
% The gestapo and german society: Political denunciation in the gestapo
%  case files.
%{\em The Journal of Modern History}~{\bf 60}, 654--694.
%
%\bibitem[Noakes and Pridham(2001)]{Noakes}
%Noakes, J. and G.~Pridham (2001).
% {\em Nazism, 1919-1945. Vol. 3: Foreign Policy, War and Racial
%  Extermination}.
%University of Exeter Press,  Exeter.
%
%\end{thebibliography}

%%%%%%%%%%%%%%%%%%%%%%%%%%%%%%%%%%%%%%%%%%%%%%%%%%%%%%%%%%%%%%%%%%%%%%%%%%%%%%%%%%%%%%%%%%%%%%%%%%%%%%%%%%%%%%%%%%%%%%%%%%%%
\vskip .65cm
\noindent
Sumin Wang
\vskip 2pt
\noindent
School of Sciences, Hebei University of Technology
\vskip 2pt
\noindent
E-mail: wangsm088@hebut.edu.cn
\vskip 2pt

\vskip .65cm
\noindent
 Chenxian Huang
\vskip 2pt
\noindent
NITFID, LPMC $\&$ KLMDASR, School of Statistics and Data Science,   Nankai University
\vskip 2pt
\noindent
E-mail: 15954719878@163.com
\vskip 2pt

\vskip .65cm
\noindent
 Yongdao Zhou
\vskip 2pt
\noindent
NITFID, LPMC $\&$ KLMDASR, School of Statistics and Data Science,   Nankai University
\vskip 2pt
\noindent
E-mail: ydzhou@nankai.edu.cn
\vskip 2pt

\vskip .65cm
\noindent
Min-Qian Liu 
\vskip 2pt
\noindent
NITFID, LPMC $\&$ KLMDASR, School of Statistics and Data Science,   Nankai University, Tianjin 300071, China
\vskip 2pt
\noindent
E-mail:  mqliu@nankai.edu.cn

% \vskip .3cm
%\centerline{(Received ???? 20??; accepted ???? 20??)}\par
\end{document}